\documentclass[11pt, a4paper]{lumia}

\usepackage[sort&compress]{natbib}
\usepackage{fontawesome5}
\usepackage{mathtools}
\usepackage{algorithm}
\usepackage{algorithmic}
\usepackage{listings}
\usepackage{multirow}
\usepackage{makecell}
\usepackage{subcaption}
\usepackage{wrapfig}
\usepackage[normalem]{ulem}
\usepackage{tikz}
\usepackage{fancyvrb}
\usepackage{framed}
\usepackage{array}
\usepackage{xspace}
\usepackage{placeins}
\usepackage{bm}

\definecolor{questionborder}{RGB}{126,174,220}
\definecolor{questionbg}{RGB}{247,250,253}

\newtcolorbox{keyquestion}{
  width=\columnwidth,
  colback=questionbg,
  colframe=questionborder,
  boxrule=0.8pt,
  arc=2mm,
  left=5pt,
  right=5pt,
  top=4pt,
  bottom=4pt,
  before skip=6pt,
  after skip=6pt,
  fontupper=\itshape,
  enhanced
}

\definecolor{prompttitlegray}{RGB}{232,232,232}
\definecolor{rulegray}{RGB}{90,90,90}
\definecolor{caseblue}{RGB}{38,65,93}
\definecolor{casegray}{RGB}{247,248,250}
\definecolor{casegrayline}{RGB}{145,151,158}
\definecolor{casegreen}{RGB}{242,249,237}
\definecolor{casegreenline}{RGB}{174,207,151}
\definecolor{caseblueback}{RGB}{239,246,250}
\definecolor{promptnavy}{RGB}{30,40,60}
\definecolor{promptcontentbg}{RGB}{250,250,252}

\newcolumntype{Y}{>{\raggedright\arraybackslash}X}
\newcolumntype{C}{>{\centering\arraybackslash}X}

\newcounter{promptbox}[section]
\renewcommand{\thepromptbox}{\thesection.\arabic{promptbox}}
\lstdefinestyle{PromptTextStyle}{%
  basicstyle=\ttfamily\footnotesize,
  breaklines=true,
  breakatwhitespace=false,
  breakautoindent=false,
  columns=flexible,
  keepspaces=true,
  xleftmargin=0pt,
  xrightmargin=0pt,
  aboveskip=0pt,
  belowskip=0pt
}
\newtcblisting{promptbox}[1]{
  enhanced,
  breakable,
  colback=promptcontentbg,
  colframe=promptnavy,
  boxrule=1pt,
  arc=2mm,
  width=0.92\textwidth,
  center,
  left=8pt, right=8pt, top=4pt, bottom=6pt,
  before skip=6pt, after skip=6pt,
  title={\small\bfseries\sffamily Prompt~\thepromptbox: #1},
  colbacktitle=promptnavy,
  coltitle=white,
  fonttitle=\small\bfseries\sffamily,
  boxed title style={boxrule=0pt,colback=promptnavy,arc=2mm},
  attach boxed title to top left={xshift=0pt,yshift=0pt},
  listing only,
  listing options={style=PromptTextStyle}
}

\newtcolorbox{caseflowbox}[1]{%
  enhanced,
  colback=white,
  colframe=caseblue,
  boxrule=0.9pt,
  arc=3mm,
  left=3.5mm,right=3.5mm,top=2.2mm,bottom=2.2mm,
  title={#1},
  colbacktitle=white,
  coltitle=caseblue,
  fonttitle=\large\bfseries,
  boxed title style={boxrule=0pt,colback=white},
  attach boxed title to top left={xshift=2mm,yshift=-1.4mm},
  before skip=2mm,after skip=1mm}

\newtcolorbox{casepromptbox}[1]{%
  enhanced,
  colback=casegray,
  colframe=casegrayline,
  boxrule=0.55pt,
  arc=2mm,
  left=2.5mm,right=2.5mm,top=1.5mm,bottom=1.5mm,
  title={#1},
  colbacktitle=casegray,
  coltitle=black,
  fonttitle=\bfseries\small,
  before skip=1.5mm,after skip=1.5mm}

\newtcolorbox{casevisualbox}[1]{%
  enhanced,
  colback=white,
  colframe=casegrayline,
  boxrule=0.55pt,
  arc=2mm,
  left=2.5mm,right=2.5mm,top=1.5mm,bottom=1.5mm,
  title={#1},
  colbacktitle=white,
  coltitle=black,
  fonttitle=\bfseries\small,
  before skip=1.5mm,after skip=1.5mm}

\newtcolorbox{caseresponsebox}[1]{%
  enhanced,
  colback=casegreen,
  colframe=casegreenline,
  boxrule=0.65pt,
  arc=2.5mm,
  left=3mm,right=3mm,top=1.7mm,bottom=1.7mm,
  title={#1},
  colbacktitle=casegreen,
  coltitle=black,
  fonttitle=\bfseries\small,
  before skip=1.5mm,after skip=1.5mm}

\newtcolorbox{caseresultbox}[1]{%
  enhanced,
  colback=caseblueback,
  colframe=caseblue!55,
  boxrule=0.65pt,
  arc=2.5mm,
  left=3mm,right=3mm,top=1.7mm,bottom=1.7mm,
  title={#1},
  colbacktitle=caseblueback,
  coltitle=black,
  fonttitle=\bfseries\small,
  before skip=1.5mm,after skip=1.5mm}

\setheadertext{Preprint}

\renewcommand{\today}{August 2026}

\title{VeinCast: Physics-Guided Dynamic Field Graphs with Graph-Conditioned Fusion for Global Medium-Range Weather Forecasting}
\setheadertitle{\resizebox{0.98\textwidth}{!}{VeinCast: Physics-Guided Dynamic Field Graphs with Graph-Conditioned Fusion for Global Medium-Range Weather Forecasting}}

\author{%
\resizebox{\linewidth}{!}{\strut
Zhisheng Chen\textsuperscript{\rm 1,*},
Jinhan Li\textsuperscript{\rm 1,*},
Yuxuan Li\textsuperscript{\rm 1},
Yuan Gao\textsuperscript{\rm 2},
Hao Wu\textsuperscript{\rm 2},
Zheng Lu\textsuperscript{\rm 3},
Jinlong Du\textsuperscript{\rm 4,$\dagger$},
Kun Wang\textsuperscript{\rm 1,$\dagger$},
Bo An\textsuperscript{\rm 1}}\\
{\Affilfont
\textsuperscript{\rm 1}Nanyang Technological University\quad
\textsuperscript{\rm 2}Tsinghua University\quad
\textsuperscript{\rm 3}Peking University\quad
\textsuperscript{\rm 4}Tongji University}\\
}
\correspondingemail{$*$ Equal contribution \quad $\dagger$ Corresponding authors \quad
\emailicon\ zhisheng.researcher@gmail.com
}
\githublink{https://github.com/Zhisheng-researcher/VeinCast}

\begin{document}

\begin{abstract}
Global medium-range weather forecasting requires modeling structured yet state-dependent interactions among heterogeneous atmospheric fields. Existing data-driven models largely learn these interactions implicitly, whereas equation-level physical constraints may inherit approximation and model-form biases. We present VeinCast, a physics-guided dynamic field graph and graph-conditioned fusion framework that jointly forecasts 69 surface and upper-air fields. Within each local window, its Physics-Guided Dynamic Field Graph combines predefined atmospheric relations with state-dependent Top-\(K\) residual edges and adapts Earth-window attention using the resulting graph context. Graph-Conditioned Latent Fusion further employs graph context and source-node centrality to guide field-to-latent aggregation, while bounded feedback preserves field-specific information. On the \(1.5^\circ\) ERA5 benchmark, VeinCast demonstrates competitive forecasting performance across all 69 meteorological fields at lead times of up to 14 days, compared with representative global weather forecasting models including FuXi, Pangu-Weather, GraphCast, FengWu, and ARROW. Ablations confirm that the two modules provide complementary gains, demonstrating the effectiveness of relational-level physical guidance for data-driven weather forecasting.
\end{abstract}

\maketitle

\setcounter{secnumdepth}{0}


\section{Introduction}
\label{sec:introduction}

\begin{figure}[t!]
    \centering
    \includegraphics[width=0.7\linewidth]{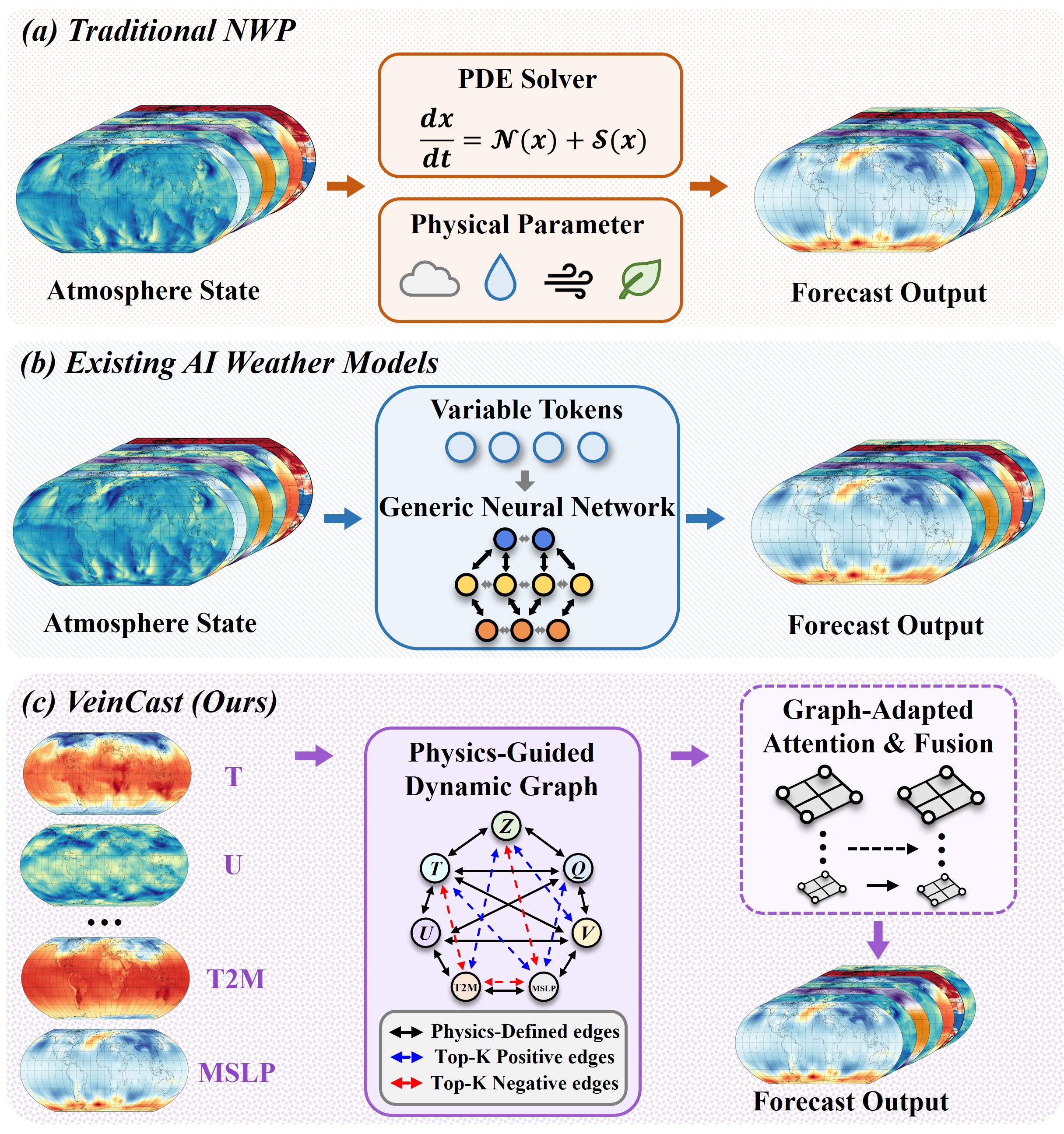}
    \caption{
    Overview of weather forecasting paradigms and VeinCast.
    \textbf{(a)} Traditional NWP relies on numerical solvers and physical parameterizations.
    \textbf{(b)} Existing AI models process atmospheric variables using generic neural networks.
    \textbf{(c)} VeinCast models structured variable interactions through a physics-guided dynamic field graph and graph-conditioned latent fusion.
    }
    \label{fig:difference}
\end{figure}

Medium-range weather forecasting predicts atmospheric evolution over lead times from several days to approximately two weeks and supports weather-sensitive decisions in disaster preparedness, agriculture, energy, transportation, and water-resource management~\cite{2025Probabilistic,innovation}. Reliable global forecasting requires jointly modeling heterogeneous surface and upper-air fields across multiple vertical levels~\cite{lianbangwea}. Their interactions exhibit persistent physical structure, but their relevance and strength vary across locations, pressure levels, and atmospheric states~\cite{al2010review}. Capturing these structured yet state-dependent cross-field couplings is therefore essential for skillful and computationally efficient medium-range forecasting.

Operational forecasting traditionally relies on Numerical Weather Prediction (NWP), which uses data assimilation to initialize the atmosphere and numerically integrates discretized governing equations~\cite{2015the}. Although physically grounded, high-resolution integration and ensemble forecasting are computationally demanding, while errors from finite resolution, unresolved-process parameterizations, numerical discretization, and imperfect initial conditions accumulate with lead time~\cite{allen2025end}. These trade-offs have motivated data-driven approaches that learn atmospheric state transitions from historical data~\cite{2010Precipitation}.

Recent AI forecasters have established a competitive data-driven paradigm for global weather forecasting while substantially reducing forecast-generation costs~\cite{ren2021deep,fourcastnet,2023Accurate,lam2023learning,chen2023fengwupushingskillfulglobal,2023FuXi}. Nevertheless, many architectures represent meteorological fields as channels or tokens and infer their interactions through generic convolution, attention, or message-passing operators~\cite{gan2025ewmoe}. Graph-based forecasters likewise construct graphs primarily over spatial grid or mesh nodes, leaving cross-field dependencies implicit in feature representations~\cite{gao2025oneforecast}. Such designs can learn statistical couplings but do not explicitly distinguish persistent atmospheric relations from interactions whose relevance changes with the local state. Vertical dependencies and couplings among wind, geopotential, temperature, and moisture have recognizable physical structure, yet their effective strengths vary across atmospheric conditions.

Physics--AI hybrid models offer another route by embedding governing equations, differentiable solvers, or equation-based constraints into neural forecasting~\cite{0Neural,xu2025generalizingweatherforecastfinegrained}. Although these approaches provide strong physical guidance, the incorporated formulations may still reflect approximations, closure assumptions, parameterizations, and numerical discretization choices. Imposing such formulations too rigidly may transfer model-form biases to the learning process and restrict data-driven correction, while remaining errors can be amplified during chaotic forecast rollout. This motivates incorporating physical knowledge at the relational rather than equation level: selected relationships provide a stable inductive bias without prescribing exact field evolution. However, a fixed relation graph cannot describe all regime-dependent interactions, whereas learning all relations without guidance may overlook robust atmospheric knowledge. We therefore combine a physics-defined backbone with data-dependent connections that capture couplings emerging under specific local states. The resulting graph should also guide multiscale spatial modeling and cross-field aggregation rather than remain an isolated auxiliary representation, while avoiding excessive fusion that obscures field-specific characteristics.

To address these challenges, we propose VeinCast, a physics-guided dynamic field graph and graph-conditioned fusion framework for global medium-range weather forecasting. VeinCast learns a 6-hour state transition and generates medium-range forecasts through autoregressive rollout. We consider a closed-set setting with 69 input and output fields, including 65 upper-air fields from five variables at 13 predefined pressure levels and four surface fields. Within each local spatial window, these fields form the nodes of a directed graph. Physics-defined edges provide a stable relational scaffold, while state-dependent Top-\(K\) residual edges capture additional couplings absent from the predefined relations. The resulting graph context adapts Earth-window spatial attention and conditions field-to-latent aggregation through source-node centrality. A Latent U-Backbone models multiscale spatial dependencies and interactions among fusion latents, after which bounded feedback returns the fused context without overwhelming field-specific features. Finally, the Forecast Decoder reconstructs the fixed forecast fields as residual atmospheric updates.

In summary, our main contributions are as follows:
\begin{itemize}
    \item We introduce a Physics-Guided Dynamic Field Graph that incorporates physical knowledge at the relational rather than equation level. It combines a stable physics-defined scaffold to distinguish persistent atmospheric relations from locally varying couplings, while using the graph context to adapt Earth-window spatial attention.
    
    \item We develop Graph-Conditioned Latent Fusion, which uses dynamic graph context and source-node centrality to guide field-to-latent aggregation. A multiscale latent backbone captures shared interactions, while bounded feedback returns the fused context without obscuring field-specific information.
    
    \item Systematic experiments on global medium-range forecasting demonstrate competitive accuracy across representative surface and upper-air fields. Ablation studies further verify the complementary contributions of the dynamic field graph and graph-conditioned fusion.
\end{itemize}


\section{Related Work}
\label{sec:related-work}

\subsection{Numerical Weather Prediction.}
NWP initializes the atmosphere through data assimilation and advances it by numerically solving discretized governing equations~\cite{ecmwf}. Although physically grounded, high-resolution and ensemble forecasting remain computationally expensive and are affected by initial-condition, resolution, parameterization, and discretization errors~\cite{78758,nguyen2023climaxfoundationmodelweather}.

\subsection{Data-Driven Weather Forecasting.}
Recent advances in deep learning and large-scale reanalysis datasets have accelerated the development of data-driven weather forecasting. Representative Transformer-based models include Pangu-Weather, FengWu, FuXi, Stormer~\cite{NEURIPS2024_7f19b99e}, STCast~\cite{stcast}, EMFormer~\cite{chen2026emformer}, and Aurora~\cite{bodnar2025foundation}. Graph-based approaches include Keisler's model~\cite{keisler2022forecastingglobalweathergraph}, GraphCast, OneForecast, and Graph-EFM~\cite{NEURIPS2024_49259289}, while diffusion-based methods include GenCast and FengWu-ENS~\cite{0The}.

Beyond purely data-driven approaches, physics--AI hybrid methods incorporate physical mechanisms into neural forecasting. NeuralGCM and ClimODE introduce differentiable dynamics or physics-informed differential equations~\cite{verma2024climodeclimateweatherforecasting}, WeatherGFT embeds governing partial differential equations, and NowcastNet combines physical evolution with conditional generation~\cite{2023Skilful}. These methods primarily inject physics at the equation or evolution level.

Existing global medium-range models generally mix fields implicitly or construct graphs over spatial grids or meshes or incorporate equation-level physical guidance. Unlike these approaches, VeinCast organizes variable--pressure-level fields as graph nodes within local spatial windows and combines a physics-defined relational backbone with state-dependent residual connections. To the best of our knowledge, this is the first such framework for global medium-range weather forecasting.


\section{Method}
\label{sec:method}

\begin{figure*}[t]
    \centering
    \includegraphics[width=\linewidth]{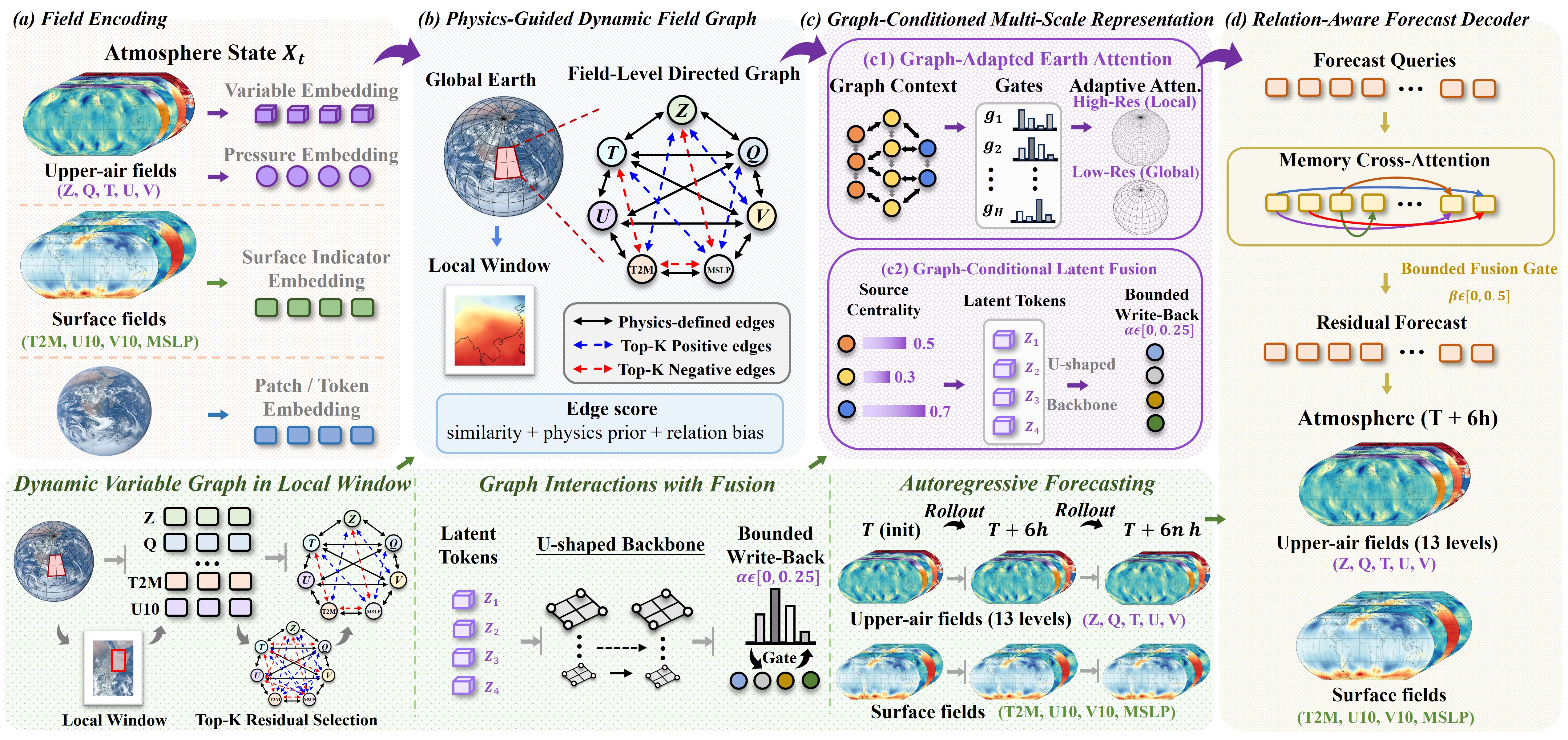}
    \caption{The overall technical flowchart of VeinCast.}
    \label{fig:veincast_overall}
\end{figure*}

Figure. \ref{fig:veincast_overall} provides an overview of VeinCast. It encodes surface and upper-air fields, constructs a Physics-Guided Dynamic Field Graph to adapt Earth attention and guide Graph-Conditioned Latent Fusion, and finally recovers the standard 69-field forecast through a Relation-Aware Forecast Decoder.

\subsection{Problem Formulation}

Given the normalized global atmospheric state $\bm{X}_t\in\mathbb{R}^{B\times M\times H\times W}$ at time $t$, VeinCast learns a state-transition model with a base interval of $\delta=6\,\mathrm{h}$:
\begin{equation}
\widehat{\bm{X}}_{t+\delta}=\mathcal{F}_{\theta}\left(\bm{X}_t,\bm{S},\bm{P}_t,\mathbf{e}^{\mathrm{cal}}_{t+\delta}\right),
\label{eq:problem}
\end{equation}
where $\bm{S}$ denotes static geographic features, $\bm{P}_t$ is a field-availability mask used to exclude invalid source fields, and $\mathbf{e}^{\mathrm{cal}}_{t+\delta}$ represents calendar conditions at the target valid time. We use $M=69$ meteorological fields, consisting of 65 upper-air fields from five variables at 13 pressure levels and four surface fields. Long-range forecasts are obtained by recursively applying the 6-hour transition model. VeinCast combines field encoding with a physics-guided dynamic field graph and graph-conditioned latent fusion, followed by a relation-aware decoder for the registered 69 forecast fields.

\subsection{Field Encoding}

Surface and upper-air fields have different vertical semantics and are therefore processed by two lightweight stems with independent parameters. Each scalar field is projected into spatial patch tokens and augmented with field metadata:
\begin{equation}
\bm{e}_m=\bm{E}_{\mathrm{var}}(v_m)+\bm{E}_{\mathrm{pres}}(\widetilde p_m,s_m)+\bm{E}_{\mathrm{surf}}(s_m),
\label{eq:field_embedding}
\end{equation}
where $v_m$ is the variable identity, $s_m\in\{0,1\}$ is the surface indicator, and $\widetilde p_m=\log(p_m/1000)$ for upper-air fields while $\widetilde p_m=0$ for surface fields. The land--sea mask, orography, and spherical positional features are projected separately and added to all field representations. The two stems are then merged in the canonical 69-field order and processed by a shared multi-scale Earth-aware encoder.

\subsection{Physics-Guided Dynamic Field Graph}

\subsubsection{Regional Descriptors}

VeinCast constructs a directed graph within each local spatial window, where nodes represent meteorological fields rather than latitude--longitude locations. Given $\bm{X}\in\mathbb{R}^{B\times M\times h\times w\times d}$, the grid is partitioned into $R$ windows, and the $T$ spatial tokens of each field are averaged into a regional descriptor:
\begin{equation}
\bm{D}_{b,r,m}=\frac{1}{T}\sum_{\tau=1}^{T}\bm{X}^{\mathrm{win}}_{b,m,r,\tau}.
\label{eq:graph_descriptor}
\end{equation}

\subsubsection{Physical Relations}

For target field $i$ and source field $j$, the relation score combines state-dependent similarity, a relation-type bias, and a physics-derived prior:
\begin{equation}
u_{b,r,i,j}=\frac{\left(W_q\bm{D}_{b,r,i}\right)^\top\left(W_k\bm{D}_{b,r,j}\right)}{\sqrt{d}}+e_{\bar{\rho}_{ij}}+\lambda P^{\mathrm{phy}}_{ij}.
\label{eq:graph_score}
\end{equation}
Here, $P^{\mathrm{phy}}_{ij}\in\{0,1\}$ indicates whether a predefined rule permits information flow from $j$ to $i$, and $\bar{\rho}_{ij}$ is its relation type. The prior contains self, within-variable vertical, wind--geopotential, temperature--moisture, moisture-transport, and surface--atmosphere relations. Pairs without a predefined relation share a learnable generic residual type. Thus, the rules specify a stable relational support while the edge strengths remain conditioned on the local atmospheric state.

\subsubsection{Dynamic Residual Edges}

For each sample, window, and target field, the source set $\Omega_{b,r,i}$ contains all physics-defined edges and the $K$ highest-scoring residual edges among the remaining candidates; sources masked by $\bm{P}_t$ are excluded. The normalized relation weights and relation-specific context are
\begin{equation}
A_{b,r,i,j}=\operatorname{Softmax}_{j\in\Omega_{b,r,i}}\left(u_{b,r,i,j}\right),
\end{equation}
\begin{equation}
\bm{C}_{b,r,i}=\sum_{\ell=1}^{N_{\rho}}\sum_{j\in\Omega_{b,r,i}}A_{b,r,i,j}\mathbf{1}[\bar{\rho}_{ij}=\ell]W_v^\ell\bm{D}_{b,r,j},
\label{eq:graph_context}
\end{equation}
where each relation type has an independent value projection $W_v^\ell$. The physics-defined edges form the relational backbone, whereas the top-$K$ residual edges capture state-dependent dependencies not covered by the rule set.

\subsubsection{Graph-Adapted Earth Attention}

The graph context is broadcast to the tokens of its spatial window and combined with field and static features to generate an adapted key and a per-head gate:
\begin{equation}
\bm{K}^{\mathrm{ada}}=W_{\mathrm{ada}}f_{\mathrm{ada}}\left([\bm{X},\bm{C},\bm{S}]\right),
\end{equation}
\begin{equation}
\bm{g}^{\mathrm{key}}=\sigma\left(f_{\mathrm{gate}}([\bm{X},\bm{C},\bm{S}])\right),
\end{equation}
\begin{equation}
\bm{K}=\bm{g}^{\mathrm{key}}\odot\bm{K}^{\mathrm{base}}+\left(1-\bm{g}^{\mathrm{key}}\right)\odot\bm{K}^{\mathrm{ada}}.
\label{eq:graph_key}
\end{equation}
The dynamic graph therefore modulates Earth-window spatial attention without replacing its native spatial representation.

\subsection{Graph-Conditioned Latent Fusion}

\subsubsection{Graph Context and Source-Node Centrality}

At both spatial resolutions, the encoder alternates dynamic graph modules with Earth-window attention. The final low-resolution graph is also reused to condition multivariate fusion. Since $A_{b,r,i,j}$ measures information flow from source $j$ to target $i$, the source centrality of field $j$ and its latent-specific bias are defined as
\begin{equation}
c_{b,r,j}=\frac{1}{M}\sum_{i=1}^{M}A_{b,r,i,j},
\end{equation}
\begin{equation}
b^{\mathrm{cent}}_{b,r,n,j}=\left[f_{\mathrm{cent}}\left(\log\left(\max(c_{b,r,j},\epsilon)\right)\right)\right]_n,
\label{eq:centrality_bias}
\end{equation}
where $n$ indexes a fusion latent. The bias is broadcast to the corresponding spatial window. After enriching low-resolution fields with graph context, the fusion latents aggregate the available fields through
\begin{equation}
\widetilde{\bm{X}}^l=W_x[\bm{X}^l,\bm{C}^l],
\end{equation}
\begin{equation}
\bm{Z}^1=\bm{Z}^0+\operatorname{MHA}\left(\bm{Z}^0,\widetilde{\bm{X}}^l,\widetilde{\bm{X}}^l;\bm{b}^{\mathrm{cent}},\bm{P}_t\right).
\label{eq:field_to_fusion}
\end{equation}
The graph context modifies field values, while source-node centrality biases field-to-latent aggregation toward locally influential source fields.

\subsubsection{Latent U-Backbone}

The fusion latents are initialized by learnable embeddings, static geography, and calendar conditions. They are processed by a Latent U-Backbone whose blocks alternate Earth-aware spatial attention within each latent slot and self-attention across latent slots at each location. Downsampling, a high-capacity bottleneck, and skip-connected upsampling capture multi-scale spatial dependencies while preserving interactions among latent semantics.

\subsubsection{Bounded Fusion Feedback}

The fused memory is written back to the low-resolution field representations by cross-attention. Let $\Delta\bm{X}^l=\operatorname{MHA}(\bm{X}^l,\bm{Z},\bm{Z})$. A field- and location-dependent gate is computed from the field representation, fusion response, and field metadata:
\begin{equation}
\bm{\alpha}=\alpha_{\min}+(\alpha_{\max}-\alpha_{\min})\,
\sigma\left(f_{\alpha}\left([\bm{X}^l,\Delta\bm{X}^l,\bm{e}]\right)\right),
\end{equation}
\begin{equation}
\overline{\bm{X}}^l=\bm{X}^l+\bm{\alpha}\odot\Delta\bm{X}^l,
\label{eq:fusion_feedback}
\end{equation}
where $(\alpha_{\min},\alpha_{\max})=(0,0.25)$. The bounded feedback preserves field-specific information while allowing each field to recover context from the shared multivariate memory.

\subsection{Relation-Aware Forecast Decoder}

For each of the registered 69 output fields, the decoder constructs a forecast query from its variable identity, pressure metadata and surface indicator:
\begin{equation}
\begin{aligned}
\bm{q}_i&=\bm{E}_{\mathrm{var}}(v_i)+\bm{E}_{\mathrm{pres}}(\widetilde p_i,s_i)+\bm{E}_{\mathrm{surf}}(s_i).
\end{aligned}
\label{eq:query_embedding}
\end{equation}
The query is combined with the local mean of available field features and performs relation-aware cross-attention over the low-resolution field memory:
\begin{equation}
\bm{Q}^l_{\mathrm{field}}=\operatorname{MHA}\left(\bm{q}+\overline{\bm{x}},\overline{\bm{X}}^l,\overline{\bm{X}}^l;\bm{\eta},\bm{P}_t\right),
\label{eq:query_attention}
\end{equation}
where $\bm{\eta}$ is a fixed bias derived from the same registered field relations, favoring the identical field, vertically related levels, and physically related variables. The field-derived query then reads the fused latent memory and combines the two sources through a bounded gate:
\begin{equation}
\bm{Q}^l_{\mathrm{fusion}}=\operatorname{MHA}\left(\bm{Q}^l_{\mathrm{field}},\bm{Z},\bm{Z}\right),
\end{equation}
\begin{equation}
\begin{aligned}
\bm{\beta}&=\beta_{\min}+(\beta_{\max}-\beta_{\min})
&\quad\cdot\sigma\left(f_{\beta}\left([\bm{Q}^l_{\mathrm{field}},\bm{Q}^l_{\mathrm{fusion}},\bm{q}]\right)\right),
\end{aligned}
\end{equation}
\begin{equation}
\bm{Q}^l=(1-\bm{\beta})\odot\bm{Q}^l_{\mathrm{field}}+\bm{\beta}\odot\bm{Q}^l_{\mathrm{fusion}},
\label{eq:query_fusion}
\end{equation}
where $(\beta_{\min},\beta_{\max})=(0,0.5)$. Low-resolution queries are upsampled and fused with high-resolution field-attention outputs, followed by Earth-aware spatial blocks and patch recovery. For the standard 69-field forecast, the model predicts a residual update in normalized space:
\begin{equation}
\widehat{\bm{X}}_{t+\delta}=\operatorname{SC}_{L}\left(\bm{X}_t+\gamma\bm{R}_{\theta}\right),
\end{equation}
\begin{equation}
\operatorname{SC}_{L}(x)=L\tanh(x/L),
\label{eq:residual_output}
\end{equation}
where $\bm{R}_{\theta}$ is the decoded residual and $\gamma$ controls its magnitude.

\subsection{Training Objective}

VeinCast is trained only with a latitude-weighted Huber loss in normalized space. Let $e_{b,m,h,w}^{(s)}=\widehat X_{b,m,h,w}^{(s)}-X_{b,m,h,w}^{(s)}$ denote the error at rollout step $s$, and let $\omega_h=\cos(\phi_h)/\left(H^{-1}\sum_{h'=1}^{H}\cos(\phi_{h'})\right)$ be the normalized weight at latitude $\phi_h$. The loss for one forecast step is
\begin{equation}
\mathcal{L}^{(s)}=\frac{\sum_{b,m,h,w}\omega_h\,\mathcal{H}_{\kappa}\left(e_{b,m,h,w}^{(s)}\right)}{\sum_{b,m,h,w}\omega_h},
\end{equation}
\begin{equation}
\mathcal{H}_{\kappa}(e)=
\begin{cases}
\frac{1}{2}e^2,& |e|\leq\kappa,\\
\kappa\left(|e|-\frac{1}{2}\kappa\right),& |e|>\kappa,
\end{cases}
\end{equation}
\begin{equation}
\mathcal{L}_{\mathrm{train}}=\frac{1}{S}\sum_{s=1}^{S}\mathcal{L}^{(s)}.
\label{eq:training_objective}
\end{equation}
Here, $S$ is the number of supervised rollout steps. No auxiliary reconstruction term or physical regularizer is used in the reported experiments.

\section{Experiments}
\label{sec:experiments}

\subsection{Dataset and Implementation Details}
\begin{table}[t]
\caption{Three-stage training configuration. Steps, BS/GPU, Accum., and TF denote the rollout steps, per-GPU batch size, gradient-accumulation steps, and teacher-forcing probability, respectively.}
\label{tab:training_schedule}
\centering
\small
\renewcommand{\arraystretch}{1.0}
\setlength{\tabcolsep}{1.5mm}
\begin{tabular}{c||c|c|c|c|c|c}
\toprule
\hline
Stage & Steps & Epochs & BS/GPU & Accum. & LR & TF \\
\hline
\hline
1   & 1 & 150 & 6 & 1 & \(1\times10^{-4}\) & 0 \\
2   & 2 & 10  & 6 & 4 & \(1\times10^{-5}\) & 0.25 \\
3   & 4 & 10  & 6 & 4 & \(1\times10^{-5}\) & 0.25 \\
\hline
\bottomrule
\end{tabular}
\end{table}

\paragraph{Dataset.}
We conduct experiments on the ERA5 reanalysis dataset at 6-hour intervals and a spatial resolution of \(1.5^{\circ}\) (\(121\times240\) grids) provided via WeatherBench2~\cite{78758,wb2}. Data from 1979--2017, 2018, and 2020 are used for training, validation, and testing, respectively. Each atmospheric state contains 69 input and output fields: geopotential (\(z\)), temperature (\(t\)), specific humidity (\(q\)), and zonal and meridional winds (\(u\) and \(v\)) at 13 pressure levels,
(\{300,400,500,600,700,850,925,1000\} hPa),
together with four surface variables: \(\mathrm{T2M}\), \(\mathrm{U10}\), \(\mathrm{V10}\), and mean sea-level pressure (\(\mathrm{MSL}\)). Per-field normalization statistics are computed only from the training set. VeinCast and baseline models predict a 6-hour state transition and generate longer-range forecasts autoregressively.

\paragraph{Implementation Details.}
VeinCast and baseline models are trained on 16 NVIDIA A100 GPUs. VeinCast using AdamW with a weight decay of \(10^{-3}\) and a linear warm-up followed by cosine learning-rate decay. We optimize a latitude-weighted Huber loss with threshold $\kappa=2.0$. Training follows a three-stage rollout curriculum, with each stage initialized from the best checkpoint of the preceding stage. The three-stage training configurations are shown in Table \ref{tab:training_schedule}. All baseline models are trained under a unified experimental framework, and their hyperparameters are systematically optimized to obtain the best-performing checkpoints for a fair comparison with VeinCast.

\begin{table*}[t!]
\caption{RMSE and ACC of VeinCast and baseline models on selected variables. The best results are highlighted in \textbf{bold}, and the second-best are \underline{underlined}.}
\label{tab:main_results}
\centering
\footnotesize
\renewcommand{\arraystretch}{1.0}
\setlength{\tabcolsep}{1.0mm}
\resizebox{\textwidth}{!}{%
\begin{tabular}{cc|cc|cc|cc|cc|cc|cc}
\toprule
\hline
\multicolumn{2}{c|}{Methods} & \multicolumn{2}{c|}{FengWu} & \multicolumn{2}{c|}{GraphCast} & \multicolumn{2}{c|}{Pangu-Weather} & \multicolumn{2}{c|}{ARROW} & \multicolumn{2}{c|}{FuXi} & \multicolumn{2}{c}{VeinCast} \\
\hline
Variable & Lead Time & RMSE$\downarrow$ & ACC$\uparrow$ & RMSE$\downarrow$ & ACC$\uparrow$ & RMSE$\downarrow$ & ACC$\uparrow$ & RMSE$\downarrow$ & ACC$\uparrow$ & RMSE$\downarrow$ & ACC$\uparrow$ & RMSE$\downarrow$ & ACC$\uparrow$ \\
\hline
\hline
\multirow{5}{*}{T2M} & 3-day & 3.29 & 0.985 & 3.83 & 0.968 & \underline{2.20} & \textbf{0.990} & 4.06 & 0.985 & 5.08 & 0.903 & \textbf{2.17} & \underline{0.990} \\
 & 5-day & 4.44 & 0.976 & 4.82 & 0.949 & \underline{2.94} & \underline{0.982} & 6.75 & 0.967 & 5.80 & 0.895 & \textbf{2.75} & \textbf{0.984} \\
 & 7-day & 5.28 & 0.969 & 5.59 & 0.932 & \underline{3.69} & \underline{0.974} & 9.28 & 0.943 & 6.49 & 0.886 & \textbf{3.27} & \textbf{0.977} \\
 & 10-day & 6.18 & 0.961 & 6.40 & 0.912 & \underline{4.40} & \underline{0.965} & 12.87 & 0.888 & 7.37 & 0.874 & \textbf{3.89} & \textbf{0.968} \\
 & 14-day & 6.91 & 0.954 & 6.98 & 0.896 & \underline{4.88} & \underline{0.958} & 18.36 & 0.718 & 8.33 & 0.852 & \textbf{4.47} & \textbf{0.959} \\
\hline
\multirow{5}{*}{U10} & 3-day & 2.64 & 0.878 & 2.96 & 0.861 & 2.85 & 0.860 & \underline{2.34} & \textbf{0.910} & 2.93 & 0.822 & \textbf{2.31} & \underline{0.908} \\
 & 5-day & 3.65 & 0.765 & 4.42 & 0.702 & 3.66 & 0.767 & \underline{3.57} & \underline{0.794} & 3.88 & 0.728 & \textbf{3.29} & \textbf{0.816} \\
 & 7-day & 4.32 & 0.668 & 5.47 & 0.561 & \underline{4.27} & \underline{0.682} & 4.58 & 0.664 & 4.61 & 0.634 & \textbf{4.07} & \textbf{0.720} \\
 & 10-day & 4.85 & 0.578 & 6.26 & 0.442 & \underline{4.81} & \underline{0.598} & 5.57 & 0.515 & 5.24 & 0.540 & \textbf{4.73} & \textbf{0.622} \\
 & 14-day & 5.15 & 0.523 & 6.66 & 0.382 & \underline{5.13} & \underline{0.548} & 6.51 & 0.383 & 5.61 & 0.476 & \textbf{5.10} & \textbf{0.554} \\
\hline
\multirow{5}{*}{V10} & 3-day & 2.70 & 0.809 & 2.88 & 0.796 & 2.90 & 0.779 & \underline{2.38} & \underline{0.855} & 2.87 & 0.772 & \textbf{2.36} & \textbf{0.862} \\
 & 5-day & 3.75 & 0.625 & 4.34 & 0.557 & 3.76 & 0.622 & \underline{3.63} & \underline{0.679} & 3.86 & 0.614 & \textbf{3.42} & \textbf{0.698} \\
 & 7-day & 4.49 & 0.461 & 5.35 & 0.345 & \underline{4.43} & 0.473 & 4.66 & \underline{0.481} & 4.65 & 0.456 & \textbf{4.26} & \textbf{0.530} \\
 & 10-day & 5.05 & 0.323 & 6.12 & 0.163 & \underline{5.01} & \underline{0.331} & 5.61 & 0.281 & 5.28 & 0.305 & \textbf{4.95} & \textbf{0.366} \\
 & 14-day & 5.31 & 0.256 & 6.40 & 0.089 & \textbf{5.28} & \underline{0.259} & 6.39 & 0.161 & 5.63 & 0.222 & \underline{5.31} & \textbf{0.271} \\
\hline
\multirow{5}{*}{Q500} & 3-day & 0.68 & 0.855 & 0.69 & 0.867 & 0.68 & 0.856 & \underline{0.64} & \underline{0.874} & 0.73 & 0.805 & \textbf{0.60} & \textbf{0.885} \\
 & 5-day & 0.85 & 0.775 & 0.94 & 0.758 & 0.86 & 0.762 & \underline{0.83} & \underline{0.786} & 0.88 & 0.738 & \textbf{0.75} & \textbf{0.824} \\
 & 7-day & \underline{0.96} & \underline{0.712} & 1.13 & 0.657 & 0.97 & 0.691 & 0.99 & 0.700 & 0.99 & 0.678 & \textbf{0.86} & \textbf{0.768} \\
 & 10-day & \underline{1.05} & \underline{0.650} & 1.29 & 0.554 & 1.05 & 0.628 & 1.18 & 0.589 & 1.10 & 0.611 & \textbf{0.97} & \textbf{0.701} \\
 & 14-day & 1.12 & \underline{0.603} & 1.40 & 0.473 & \underline{1.10} & 0.588 & 1.43 & 0.454 & 1.19 & 0.551 & \textbf{1.06} & \textbf{0.642} \\
\hline
\multirow{5}{*}{T850} & 3-day & 2.32 & 0.980 & 2.34 & 0.981 & \underline{2.10} & \underline{0.986} & 2.17 & 0.985 & 3.86 & 0.903 & \textbf{1.85} & \textbf{0.987} \\
 & 5-day & 3.28 & 0.960 & 3.66 & 0.954 & \underline{3.05} & \underline{0.970} & 3.11 & 0.967 & 4.67 & 0.889 & \textbf{2.56} & \textbf{0.975} \\
 & 7-day & 4.07 & 0.940 & 4.68 & 0.926 & \underline{3.89} & \underline{0.952} & 4.08 & 0.941 & 5.37 & 0.871 & \textbf{3.30} & \textbf{0.958} \\
 & 10-day & 4.84 & 0.917 & 5.63 & 0.894 & \underline{4.72} & \underline{0.932} & 5.42 & 0.893 & 6.15 & 0.848 & \textbf{4.07} & \textbf{0.935} \\
 & 14-day & 5.42 & 0.899 & 6.21 & 0.872 & \underline{5.32} & \underline{0.916} & 7.52 & 0.807 & 6.76 & 0.826 & \textbf{4.59} & \textbf{0.918} \\
\hline
\multirow{5}{*}{Z700} & 3-day & 310.70 & 0.986 & 284.32 & 0.987 & 265.80 & 0.989 & \underline{249.80} & \underline{0.992} & 514.46 & 0.910 & \textbf{209.80} & \textbf{0.993} \\
 & 5-day & 502.52 & 0.964 & 515.32 & 0.958 & \underline{436.67} & 0.969 & 438.87 & \underline{0.974} & 685.44 & 0.893 & \textbf{388.12} & \textbf{0.976} \\
 & 7-day & 660.34 & 0.937 & 732.71 & 0.916 & \underline{580.83} & \underline{0.945} & 633.77 & 0.943 & 842.80 & 0.867 & \textbf{559.86} & \textbf{0.950} \\
 & 10-day & 809.71 & 0.906 & 966.41 & 0.865 & \textbf{723.39} & \underline{0.914} & 872.87 & 0.890 & 1007.41 & 0.833 & \underline{733.77} & \textbf{0.914} \\
 & 14-day & 905.62 & 0.885 & 1133.90 & 0.834 & \textbf{812.65} & \textbf{0.891} & 1127.74 & 0.823 & 1123.30 & 0.804 & \underline{847.07} & \underline{0.885} \\
\hline
\bottomrule
\end{tabular}
}
\end{table*}

\subsection{Model Performance and Analysis}

Table~\ref{tab:main_results} compares VeinCast with five representative global weather forecasting models, including FengWu, GraphCast, Pangu-Weather, ARROW~\cite{tian2026arrow}, and FuXi. We report results for three surface variables (\(\mathrm{T2M}\), \(\mathrm{U10}\), and \(\mathrm{V10}\)) and three upper-air fields (\(\mathrm{Q500}\), \(\mathrm{T850}\), and \(\mathrm{Z700}\)) at lead times from 3 to 14 days. Performance is evaluated using latitude-weighted root mean square error (RMSE) and anomaly correlation coefficient (ACC), where lower RMSE and higher ACC indicate better forecast accuracy.

VeinCast achieves the lowest RMSE for most variable--lead-time combinations while maintaining consistently high ACC at extended forecast horizons. Across all five lead times, it delivers the best overall performance for $\mathrm{T2M}$, $\mathrm{Q500}$, and $\mathrm{T850}$, with particularly clear advantages for temperature and moisture prediction. It also performs strongly on near-surface winds, achieving the highest $\mathrm{V10}$ ACC at every lead time and the highest $\mathrm{U10}$ ACC from 5 to 14 days. These results highlight the benefit of explicitly modeling cross-field relationships and their joint temporal evolution. Moreover, the performance gains remain stable as the forecast horizon increases, indicating improved robustness against error accumulation during recursive forecasting. Although Pangu-Weather retains a modest advantage in $\mathrm{Z700}$ RMSE at 10 and 14 days, VeinCast provides stronger and more consistent performance across heterogeneous surface and upper-air variables.

\begin{figure}[t]
    \centering
    \includegraphics[width=\linewidth]{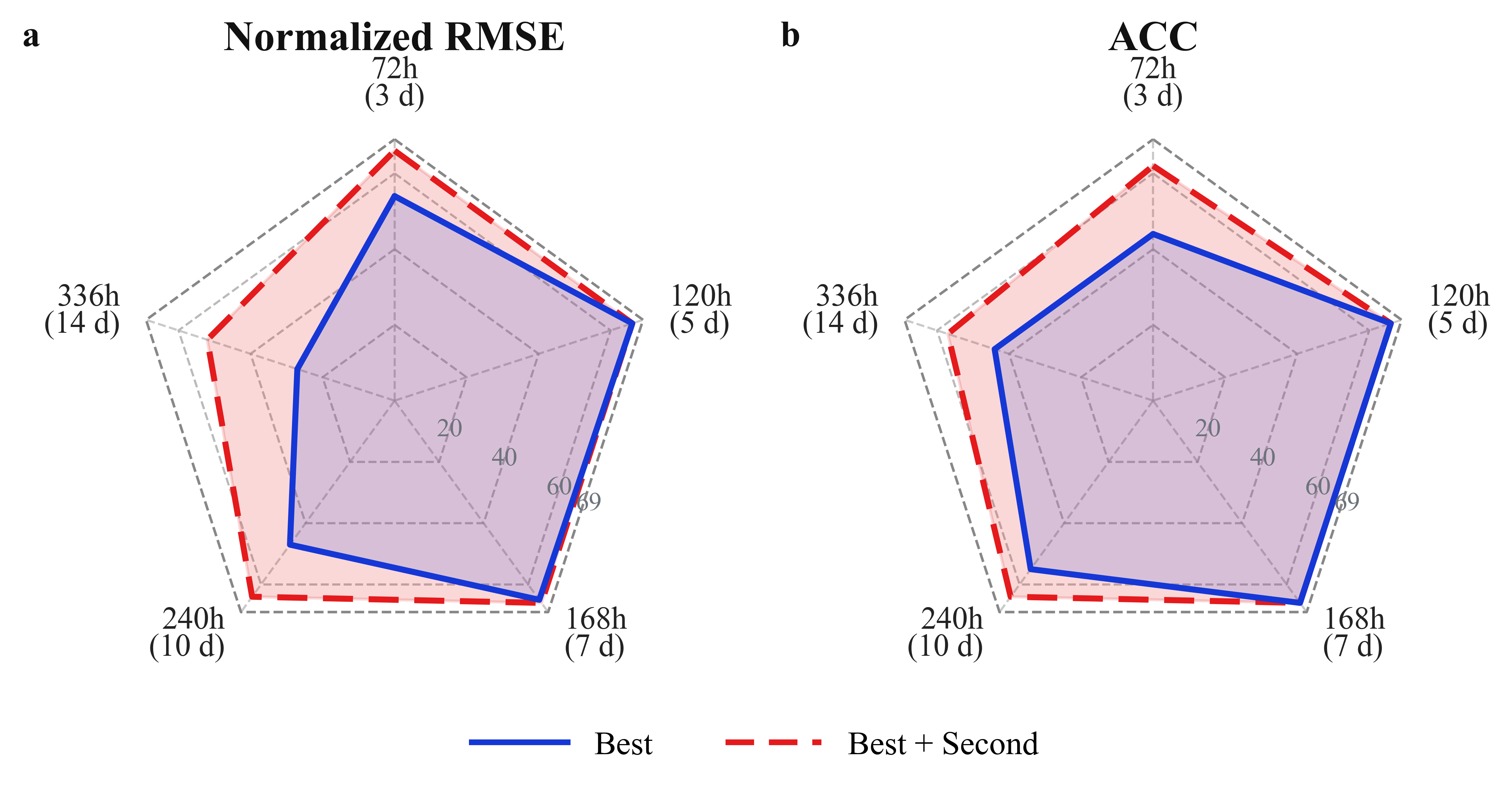}
    \caption{Rank statistics of VeinCast on normalized RMSE and ACC across different forecast lead times. Each radar chart summarizes one evaluation metric, with spokes denoting forecast lead times. “Best” denotes the number of variables for which VeinCast ranked first among all compared methods, and “Best + Second” denotes the cumulative number of variables for which VeinCast ranked first or second.}
    \label{fig:veincast_rank}
\end{figure}

Fig. \ref{fig:veincast_rank} provides a rank-based summary of VeinCast’s performance on normalized RMSE and ACC across all forecast lead times and variables. VeinCast consistently achieves the best or second-best rank for most variables, with especially strong dominance at 120 h and 168 h, indicating superior and stable performance across the full variable set.

\subsection{Ablation Study}

We conduct ablation studies to examine the contributions of the dynamic
field graph, graph-conditioned latent fusion, adaptive fusion gates, and
model capacity. Table~\ref{tab:ablation} summarizes the performance
averaged over all evaluated lead times, together with the long-range
results at lead times of at least 168 hours. The complete VeinCast
consistently provides strong overall and long-range forecasting
performance.

\begin{table}[t]
\caption{Ablation study on key components of VeinCast. N-RMSE denotes normalized RMSE; GC denotes graph-conditioned.}
\label{tab:ablation}
\centering
\small
\renewcommand{\arraystretch}{1.05}
\setlength{\tabcolsep}{5.5pt}
\begin{tabular}{lcccc}
\toprule
& \multicolumn{2}{c}{Average} & \multicolumn{2}{c}{$\geq 168$ h} \\
\cmidrule(lr){2-3}\cmidrule(lr){4-5}
Methods & N-RMSE$\downarrow$ & ACC$\uparrow$ & N-RMSE$\downarrow$ & ACC$\uparrow$ \\
\midrule
VeinCast & \textbf{0.59} & \textbf{0.807} & \textbf{0.81} & \textbf{0.704} \\
w/o Dyn. Graph & 0.69 & 0.778 & 0.96 & 0.668 \\
Static Phys. Graph & 0.64 & 0.787 & 0.88 & 0.681 \\
w/o GC Fusion & 0.69 & 0.779 & 0.95 & 0.672 \\
Fixed $\alpha/\beta$ & 0.60 & 0.802 & 0.83 & 0.698 \\
Wide Baseline & 0.69 & 0.754 & 0.94 & 0.638 \\
\bottomrule
\end{tabular}
\end{table}

\textbf{Dynamic field graph.}
Removing the dynamic field graph leads to a pronounced degradation in
both normalized RMSE and ACC, with the performance gap becoming more
evident at longer lead times. Retaining only the predefined physical
relations recovers a substantial portion of this loss, demonstrating
that the physical relational scaffold provides an effective structural
prior. Nevertheless, the remaining gap to the complete model indicates
that state-dependent residual edges complement the fixed physical graph
by capturing cross-field interactions that vary with the local
atmospheric state.

\textbf{Graph-conditioned fusion.}
Disabling graph conditioning in the latent fusion module also causes a
clear performance drop. In this variant, the dynamic graph remains
available to the encoder, but its contextual representation and
centrality information are not used to guide cross-field aggregation.
The resulting degradation shows that relational information is most
effective when it directly conditions the fusion and feedback processes,
rather than remaining isolated within the graph-enhanced encoder.

\textbf{Adaptive fusion gates.}
Replacing the adaptive $\alpha$ and $\beta$ gates with fixed
coefficients produces performance close to that of the complete model.
This relatively small difference suggests that adaptive gating is not
the primary source of the overall improvement. Instead, it serves as a
fine-grained mechanism for regulating field-specific feedback and the
use of fusion memory under different atmospheric states and forecast
lead times.

\textbf{Model capacity.}
The capacity-controlled Wide Baseline increases the encoder width while
removing both the dynamic graph and the latent fusion pathway, resulting
in a parameter count comparable to that of VeinCast. Despite this
increased capacity, it remains substantially inferior, particularly in
terms of long-range ACC. This result confirms that the improvements of
VeinCast arise from its relational modeling and graph-conditioned fusion
mechanisms rather than from a simple increase in model size.

\begin{figure*}[t]
    \centering
    \includegraphics[width=\textwidth]{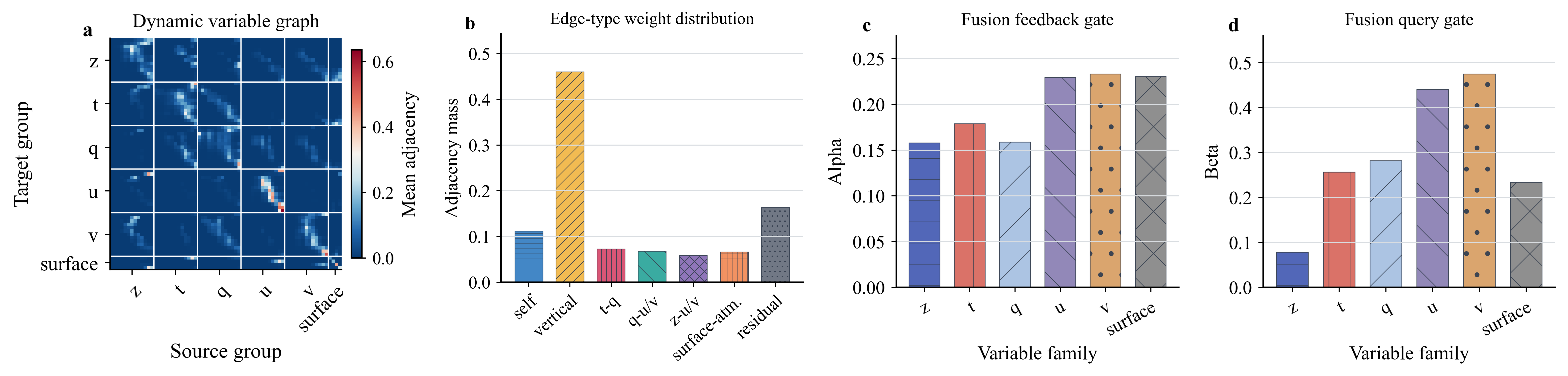}
    \caption{
    Analysis of the learned graph structure and fusion gates.
    \textbf{(a)} Learned adjacency matrix grouped by variable family, where rows and columns denote target and source groups, respectively.
    \textbf{(b)} Adjacency mass assigned to different edge types.
    \textbf{(c)} Fusion feedback gate $\alpha$ for different variable families.
    \textbf{(d)} Fusion query gate $\beta$ for different variable families.
    }
    \label{fig:mechanism}
\end{figure*}

\begin{table}[t]
\caption{Ablation study on VeinCast training stages. N-RMSE denotes normalized RMSE.}
\label{tab:ablation_stages}
\centering
\small
\renewcommand{\arraystretch}{1.05}
\setlength{\tabcolsep}{5.5pt}
\begin{tabular}{lcccc}
\toprule
& \multicolumn{2}{c}{Average} & \multicolumn{2}{c}{$\geq 168$ h} \\
\cmidrule(lr){2-3}\cmidrule(lr){4-5}
Methods & N-RMSE$\downarrow$ & ACC$\uparrow$ & N-RMSE$\downarrow$ & ACC$\uparrow$ \\
\midrule
VeinCast (Stage 1) & 0.64 & 0.785 & 0.86 & 0.678 \\
VeinCast (Stage 2) & 0.63 & 0.794 & 0.84 & 0.688 \\
VeinCast (Stage 3) & \textbf{0.59} & \textbf{0.807} & \textbf{0.81} & \textbf{0.704} \\
\bottomrule
\end{tabular}
\end{table}

\textbf{Progressive rollout training.}
Table~\ref{tab:ablation_stages} shows that the rollout fine-tuning in
Stages~2 and~3 consistently improves forecasting performance over the
one-step model. By progressively training the model on multi-step
predictions, these stages better align the training and inference
processes and mitigate error accumulation during recursive forecasting.
The results therefore confirm the effectiveness of progressive rollout
fine-tuning, particularly for improving the stability of long-range
forecasts.

\subsection{Analysis of Learned Graph Structure and Fusion Gates}

We analyze the learned adjacency matrix, edge-type weights, and fusion gates to characterize how the proposed modules organize cross-variable information. The results are shown in Fig.~\ref{fig:mechanism}.

\textbf{Learned field graph.}
The adjacency matrix exhibits clear block-wise structures across variable families. Strong weights concentrate within the same family and between vertically related atmospheric variables, whereas unrelated variable pairs receive weaker connections. Non-zero interactions between temperature and moisture, as well as between wind and pressure-related variables, further show that the graph captures structured cross-variable dependencies rather than uniformly mixing all fields.

\textbf{Edge-type weight distribution.}
Vertical edges account for the largest proportion of the adjacency mass, confirming that cross-level atmospheric coupling is the dominant relation in the learned graph. The learned residual receives the second-largest contribution, showing that the model supplements the predefined relations with data-driven connections. Self edges and the remaining physically defined relations retain non-zero weights, indicating that both prior structure and learned interactions contribute to graph construction.

\textbf{Fusion feedback gate.}
The feedback gate $\alpha$ assigns larger values to the horizontal wind and surface variables than to geopotential height, temperature, and moisture. This variable-dependent pattern shows that the feedback pathway modulates different atmospheric fields unequally, with stronger modulation for wind and surface-related predictions.

\textbf{Fusion query gate.}
The query gate $\beta$ exhibits stronger variable selectivity than $\alpha$. The largest gate values occur for the horizontal wind variables, whereas geopotential height receives the smallest value. The remaining variables lie between these two extremes. This ordering indicates that the query pathway adjusts the contribution of fused information according to the target variable family.

Overall, the learned graph combines structured atmospheric relations with complementary data-driven connections, while the two fusion gates apply distinct variable-dependent modulation. These results show that the proposed modules perform selective information integration across atmospheric variables.

\section{Conclusion}
This paper introduced VeinCast to model structured and state-dependent interactions among meteorological fields in global medium-range weather forecasting. The Physics-Guided Dynamic Field Graph combines predefined relations with state-dependent residual edges, while its context adapts Earth-window attention and guides field-to-latent aggregation through Graph-Conditioned Latent Fusion. Experiments show competitive accuracy across surface and upper-air variables, particularly at extended lead times. Ablations and graph analyses demonstrate that the physical scaffold, residual connections, and graph-conditioned aggregation are complementary, and that the gains do not arise from model capacity alone. These findings support relational-level physical priors as a middle ground between implicit data-driven mixing and rigid equation-level constraints. The current evaluation is limited to deterministic forecasts at \(1.5^\circ\) resolution over a fixed 69-field set. Future work will consider higher-resolution and probabilistic forecasting.

\bibliography{aaai2027}



\clearpage
\appendix
\section*{Appendix}
\setcounter{secnumdepth}{1}

\section{Model Details for VeinCast}
\suppressfloats[t]

This section provides a complete specification of the VeinCast architecture and
its autoregressive training procedure.  VeinCast remains a closed-set forecaster: both
its input and its supervised forecast target consist of the same registered 69
meteorological fields.

\subsection{Notation and End-to-End Tensor Flow}

At analysis time $t$, the normalized atmospheric state is
$\bm{X}_t\in\mathbb{R}^{B\times M\times H\times W}$, where $B$ is the
batch size, $M=69$ is the number of registered fields, and
$(H,W)=(121,240)$ is the latitude--longitude grid.  The field registry contains
five upper-air variables, geopotential $z$, temperature $t$, specific humidity
$q$, zonal wind $u$, and meridional wind $v$, at the 13 pressure levels
\begin{equation}
\begin{aligned}
\mathcal{P}=\{&
50,100,150,200,250,300,400,\\
&500,600,700,850,925,1000\}\ \mathrm{hPa}.
\end{aligned}
\label{eq:app_pressure_levels}
\end{equation}
together with the four surface variables T2M, U10, V10, and MSL.  We index a
field by $m\in\{1,\ldots,M\}$ and associate it with a variable identity $v_m$,
a pressure $p_m$, and a surface indicator $s_m\in\{0,1\}$.

VeinCast also receives static geographic features
$\bm{S}\in\mathbb{R}^{B\times C_s\times H\times W}$, a field-availability
mask $\bm{P}_t\in\{0,1\}^{B\times M}$, and calendar features at the forecast
valid time.  The availability mask acts on source fields throughout the graph,
fusion, and decoder modules.  A missing source therefore cannot transmit
information to another field, although the tensor shape remains fixed.

Table~\ref{tab:app_tensor_shapes} summarizes the main representations in the
default configuration.  The field dimension and the latent-slot dimension are
kept explicit throughout the network; they are never folded into ordinary
feature channels.

\begin{table}[t]
\caption{Tensor shapes in the default VeinCast configuration.  Multiplicity
denotes the explicit field or latent dimension.}
\label{tab:app_tensor_shapes}
\centering
\footnotesize
\setlength{\tabcolsep}{1.0mm}
\begin{tabular}{lccc}
\toprule
Representation & Spatial grid & Width & Multiplicity \\
\midrule
Input state & $121\times240$ & 1 & 69 fields \\
High-resolution fields & $31\times60$ & 96 & 69 fields \\
Low-resolution fields & $16\times30$ & 192 & 69 fields \\
Fusion latents & $16\times30$ & 384 & 4 slots \\
Fusion bottleneck & $8\times15$ & 768 & 4 slots \\
Decoded queries & $31\times60$ & 96 & 69 queries \\
Forecast state & $121\times240$ & 1 & 69 fields \\
\bottomrule
\end{tabular}
\end{table}

The full one-step computation can be written schematically as
\begin{equation}
\begin{aligned}
(\bm{X}^{h},\bm{X}^{l},\mathcal{G}^{l})
    &=\mathcal{E}(\bm{X}_t,\bm{S},\bm{P}_t),\\
\bm{Z}
    &=\mathcal{U}_{\mathrm{latent}}
      \left(\mathcal{A}_{F\rightarrow Z}
      (\bm{X}^{l},\mathcal{G}^{l})\right),\\
\overline{\bm{X}}^{l}
    &=\mathcal{A}_{Z\rightarrow F}(\bm{X}^{l},\bm{Z}),\\
\bm{R}_{\theta}
    &=\mathcal{D}(\bm{X}^{h},\overline{\bm{X}}^{l},\bm{Z},\bm{P}_t),\\
\widehat{\bm{X}}_{t+\delta}
    &=\operatorname{SC}_{12}
      (\bm{X}_t+0.25\bm{R}_{\theta}),
\end{aligned}
\label{eq:app_overview}
\end{equation}
where $\delta=6\,\mathrm{h}$, $\mathcal{E}$ is the two-scale field encoder,
$\mathcal{G}^{l}$ is the final low-resolution dynamic graph state,
$\mathcal{U}_{\mathrm{latent}}$ is the Latent U-Backbone, and $\mathcal{D}$ is
the relation-aware decoder.  Long-range forecasts are obtained by recursively
applying this shared 6-hour transition operator.

\subsection{Dual-Modality Field Encoding}

\subsubsection{Continuous Field Metadata}

For an upper-air field, pressure is represented on a logarithmic coordinate;
surface fields use a distinguished coordinate and an explicit surface flag:
\begin{equation}
\widetilde p_m=
\begin{cases}
\log(p_m/1000), & s_m=0,\\
0, & s_m=1.
\end{cases}
\label{eq:app_pressure_coordinate}
\end{equation}
Let $\{\omega_k\}_{k=1}^{K_p}$ be logarithmically spaced Fourier
frequencies.  The continuous pressure features are
\begin{equation}
\bm{\psi}_p(\widetilde p_m,s_m)=
\left[
\widetilde p_m,
s_m,
\{\sin(\omega_k\widetilde p_m),
\cos(\omega_k\widetilde p_m)\}_{k=1}^{K_p}
\right].
\label{eq:app_pressure_fourier}
\end{equation}
An MLP maps these features to the token width. We define
\[
E_{\mathrm{pres}}(\tilde{p}_m,s_m)
:=
f_{\mathrm{pres}}
\bigl(\psi_p(\tilde{p}_m,s_m)\bigr).
\]
The complete field metadata is
\begin{equation}
e_m
=
E_{\mathrm{var}}(v_m)
+
E_{\mathrm{pres}}(\tilde{p}_m,s_m)
+
E_{\mathrm{surf}}(s_m).
\end{equation}
The same parameterization is reused to describe output fields in the forecast
decoder.  In the reported closed-set experiments, all output queries correspond
to the registered pressure levels and surface variables.

\subsubsection{Surface and Upper-Air Stems}

Surface and upper-air fields are processed by two independently parameterized
stems.
For either modality, each scalar field is first zero-padded to a multiple of the
$4\times4$ patch size and projected by a strided convolution.  A lightweight
refinement block then applies group normalization, a depthwise $3\times3$
convolution, GELU activation, and a pointwise convolution with a residual
connection.  Longitude is padded periodically in the refinement convolution,
whereas latitude is padded by boundary replication.  The output of the stem for
field $m$ is
\begin{equation}
\begin{aligned}
\bm{X}^{h,0}_m
&=\operatorname{LN}\big(
\operatorname{Stem}_{s_m}(\bm{X}_{t,m})+\bm{e}_m+\bm{m}_m\big),
\end{aligned}
\label{eq:app_stem}
\end{equation}
where $\bm{m}_m$ is a learnable missing-field token when $P_{t,m}=0$ and zero
otherwise.  The physical input values of an unavailable field are set to zero
before the stem.  The two modality-specific outputs are then restored to the
canonical 69-field order.

Static inputs consist of the land--sea mask, orography, and spherical location
features derived from latitude $\phi$ and longitude $\lambda$.  They are
independently patch-embedded and added to every field token.  This produces
$\bm{X}^{h,0}\in\mathbb{R}^{B\times M\times31\times60\times96}$.

\subsubsection{Two-Scale Earth-Aware Encoder}

At the high-resolution scale, two Earth-window blocks alternate non-shifted and
shifted $4\times8$ windows with four attention heads.  Before every block, a
dynamic graph module is reconstructed from the current field representations and
provides the graph-adapted keys described below.  Patch merging concatenates
neighboring $2\times2$ tokens, applies layer normalization, and projects their
combined width from $4\times96$ to 192.  The resulting low-resolution grid is
$16\times30$.  Static tokens are bilinearly aligned to this grid, linearly
projected, and added again.  Four additional Earth-window blocks operate at this
scale with eight heads and alternating window shifts.

For each Earth-window block, the positional bias contains a relative-position
term shared across windows and a latitude-window-specific term:
\begin{equation}
\bm{B}^{\mathrm{earth}}_{r}
=\bm{B}^{\mathrm{rel}}+\bm{B}^{\mathrm{lat}}_{r_{\phi}},
\label{eq:app_earth_bias}
\end{equation}
where $r_{\phi}$ identifies the latitude band of window $r$.  Periodic
longitude padding respects global wraparound, while replicated latitude padding
avoids introducing an artificial connection between the two poles.

\subsection{Physics-Guided Dynamic Field Graph}

\subsubsection{Window-Level Field Descriptors}

The graph is reconstructed independently for each sample and spatial window.
Its nodes are meteorological fields, not geographic grid cells.  Given field
tokens $\bm{X}\in\mathbb{R}^{B\times M\times h\times w\times d}$, we first
pad the spatial grid to a multiple of the window size and partition it into $R$
windows containing $T$ tokens.  The regional descriptor of field $m$ is
\begin{equation}
\bm{D}_{b,r,m}=\frac{1}{T}\sum_{\tau=1}^{T}
\bm{X}^{\mathrm{win}}_{b,m,r,\tau}.
\label{eq:app_descriptor}
\end{equation}
The same descriptor is used to score candidate edges and to construct the
relation-specific graph context.  This compression avoids applying an
$M\times M$ relation operator separately at every spatial token.

\subsubsection{Exact Relation Registry}

For a directed edge $j\rightarrow i$, field $j$ is the source from which target
field $i$ reads information.  The physical registry uses the rules in
Table~\ref{tab:app_relations}.  Two upper-air fields are regarded as being at
nearby pressure levels when
\begin{equation}
\left|\log p_i-\log p_j\right|<0.45.
\label{eq:app_near_pressure}
\end{equation}
The relation rules define candidate support rather than a numerical evolution
equation: every retained edge still receives a state-dependent learned weight.

\begin{table*}[t]
\caption{Physical relation registry for the 69-field configuration.  The
surface--atmosphere mappings are T2M with $\{t,q\}$, U10 with $\{u,z\}$, V10
with $\{v,z\}$, and MSL with $\{z,u,v,t\}$.  Each valid cross-field pair is
registered in both information-flow directions.}
\label{tab:app_relations}
\centering
\small
\setlength{\tabcolsep}{2.4mm}
\begin{tabular}{llr}
\toprule
Relation type & Rule for source and target fields & Directed edges \\
\midrule
Self & Identical variable and level or identical surface field & 69 \\
Vertical & Same upper-air variable at two distinct pressure levels & 780 \\
Wind--geopotential & $z$ paired with $u$ or $v$ at nearby pressure levels & 204 \\
Temperature--moisture & $t$ paired with $q$ at nearby pressure levels & 102 \\
Moisture transport & $q$ paired with $u$ or $v$ at nearby pressure levels & 204 \\
Surface--atmosphere & Registered surface/upper-air pair at $p\geq700$ hPa & 80 \\
\bottomrule
\end{tabular}
\end{table*}

Let $\rho_{ij}\in\{1,\ldots,N_{\rho}\}$ denote a physical relation type and
$P^{\mathrm{phy}}_{ij}$ indicate whether a rule permits $j\rightarrow i$.
Pairs not covered by the registry are assigned a shared learned-residual type
for scoring.  After descriptor normalization, the edge score is
\begin{equation}
\begin{aligned}
u_{b,r,i,j}
&=d^{-1/2}
\left(W_q\operatorname{LN}(\bm{D}_{b,r,i})\right)^{\top}
&\quad\cdot
\left(W_k\operatorname{LN}(\bm{D}_{b,r,j})\right)
+e_{\bar\rho_{ij}}+\lambda P^{\mathrm{phy}}_{ij},
\end{aligned}
\label{eq:app_graph_score}
\end{equation}
where $e_{\bar\rho_{ij}}$ is a learned scalar bias and $\lambda=1$ in the
default configuration.  The relation prior raises the initial preference for a
registered physical edge without fixing its final attention weight.

\subsubsection{Residual Edge Selection and Normalization}

Source availability is applied before residual selection:
\begin{equation}
u_{b,r,i,j}\leftarrow-\infty
\quad\text{if}\quad P_{t,b,j}=0.
\label{eq:app_source_mask}
\end{equation}
For every $(b,r,i)$, all available physics-defined edges are retained.  Among
the remaining candidates, the $K=4$ highest-scoring edges are selected as
learned residuals.  With $\Omega_{b,r,i}$ denoting the union of these two sets,
the normalized adjacency is
\begin{equation}
A_{b,r,i,j}=
\frac{\exp(u_{b,r,i,j})}
{\sum_{k\in\Omega_{b,r,i}}\exp(u_{b,r,i,k})},
\qquad j\in\Omega_{b,r,i},
\label{eq:app_adjacency}
\end{equation}
and zero otherwise.  Top-$K$ selection is thus local to the sample, spatial
window, and target field.  A residual edge may change across locations or
forecast states even though the physical scaffold remains fixed.

Each relation type has an independent value projection.  Let
$\widetilde\rho_{b,r,i,j}$ be the final edge type, with all selected nonphysical
edges sharing the learned-residual type.  The field-specific graph context is
\begin{equation}
\bm{C}_{b,r,i}=
\sum_{\ell=1}^{N_{\rho}}
\sum_{j\in\Omega_{b,r,i}}
A_{b,r,i,j}
\mathbf{1}[\widetilde\rho_{b,r,i,j}=\ell]
W_v^{\ell}\bm{D}_{b,r,j}.
\label{eq:app_relation_context}
\end{equation}
This context is broadcast to all $T$ tokens of the corresponding window.  The
graph construction has complexity $O(BRM^2d)$ for a fixed windowing scheme,
rather than applying the same pairwise field operation independently to every
spatial token.

\subsubsection{Graph-Adapted Earth-Window Attention}

The graph communicates cross-field information to a spatial block by adapting
its key representation.  At each field token, graph context and static context
are concatenated with the current representation:
\begin{equation}
\bm{H}^{\mathrm{ada}}=[\bm{X},\bm{C},\bm{S}].
\end{equation}
Two small networks generate an adapted key and one gate per attention head:
\begin{equation}
\begin{aligned}
\bm{K}^{\mathrm{ada}}
&=W_{\mathrm{ada}}f_{\mathrm{ada}}(\bm{H}^{\mathrm{ada}}),\\
\bm{g}^{\mathrm{key}}
&=\sigma\left(f_{\mathrm{gate}}(\bm{H}^{\mathrm{ada}})\right).
\end{aligned}
\label{eq:app_key_adapter}
\end{equation}
After spatial window partitioning, the key used by each attention head is
\begin{equation}
\bm{K}=\bm{g}^{\mathrm{key}}\odot\bm{K}^{\mathrm{base}}
+(1-\bm{g}^{\mathrm{key}})\odot\bm{K}^{\mathrm{ada}}.
\label{eq:app_key_mix}
\end{equation}
The corresponding query and value remain those of the native Earth-window
block.  Therefore, the graph modulates within-field spatial attention without
replacing its spatial representation: a gate near one preserves the base key,
whereas a smaller gate increases the graph-adapted contribution.

\subsection{Graph-Conditioned Latent Fusion}

\subsubsection{Conditioned Latent Initialization}

At every position of the $16\times30$ grid, VeinCast maintains $N_z=4$ latent
slots with width $d_z=384$.  Their initialization combines learnable latent
identities with static, geographic, lead-time, and calendar conditions:
\begin{equation}
\begin{aligned}
\bm{Z}^{0}
&=\bm{Z}^{\mathrm{learn}}
+E_{\mathrm{static}}(\bm{S}^{l})
+E_{\mathrm{geo}}(G_{\mathrm{geo}}^l)
&\quad+E_{\mathrm{lead}}(\delta/24)
+E_{\mathrm{cal}}(m_{t+\delta},h_{t+\delta}).
\end{aligned}
\label{eq:app_latent_init}
\end{equation}
Here $G_{\mathrm{geo}}^l$ contains normalized latitude and longitude coordinates and
multi-frequency sine/cosine features.  Month and UTC hour are encoded
cyclically.  Invalid or unavailable calendar values are replaced by a learned
unknown-time embedding.  Although $E_{\mathrm{lead}}$ is continuously
parameterized, every transition in the reported model uses the trained base
interval $\delta=6\,\mathrm{h}$.

\subsubsection{Graph Context and Source Centrality}

The final low-resolution graph state is reused rather than reconstructed solely
for fusion.  First, field tokens are enriched with their graph context:
\begin{equation}
\widetilde{\bm{X}}^{l}=
W_x\operatorname{LN}([\bm{X}^{l},\bm{C}^{l}])
\in\mathbb{R}^{B\times M\times16\times30\times384}.
\label{eq:app_enriched_field}
\end{equation}
Second, because $A_{b,r,i,j}$ measures information flow from source $j$ to
target $i$, the average outgoing contribution of source $j$ is
\begin{equation}
c_{b,r,j}=\frac{1}{M}\sum_{i=1}^{M}A_{b,r,i,j}.
\label{eq:app_centrality}
\end{equation}
After lower bounding by $\epsilon=10^{-4}$ and applying a logarithm, a small MLP
maps this scalar to one bias for each latent slot:
\begin{equation}
b^{\mathrm{cent}}_{b,r,n,j}=
\left[f_{\mathrm{cent}}
\left(\log(\max(c_{b,r,j},\epsilon))\right)\right]_n.
\label{eq:app_centrality_bias}
\end{equation}
The bias is shared across attention heads but varies with sample, location,
latent slot, and source field.

At each low-resolution spatial position, fusion latents serve as queries and
fields serve as keys and values.  For head $a$, latent slot $n$, and source
field $m$, the attention logit is
\begin{equation}
\begin{aligned}
\ell_{b,y,x,a,n,m}
&=d_a^{-1/2}\big\langle
W_q^a\operatorname{LN}(\bm{Z}^{0}_{b,n,y,x}),
&\hspace{18mm}W_k^a\widetilde{\bm{X}}^{l}_{b,m,y,x}
\big\rangle+b^{\mathrm{cent}}_{b,y,x,n,m}.
\end{aligned}
\label{eq:app_field_to_fusion_logit}
\end{equation}
Unavailable source fields are masked before normalization:
\begin{equation}
\pi_{b,y,x,a,n,m}=
\operatorname{Softmax}_{m:P_{t,b,m}=1}
(\ell_{b,y,x,a,n,m}).
\label{eq:app_field_to_fusion_weight}
\end{equation}
The multi-head output is projected and added residually to $\bm{Z}^{0}$.  Graph
context and source centrality therefore play complementary roles: graph context
changes the field representation being read, while centrality biases the
relative priority of locally influential sources.

\subsubsection{Latent U-Backbone}

The graph-conditioned latents are processed by a U-shaped hierarchy.  Three
blocks operate at $16\times30$ with width 384 and 12 heads.  Patch merging then
produces an $8\times15$ representation with width 768, followed by four
bottleneck blocks with 24 heads.  Bilinear patch expansion returns to
$16\times30$ and width 384.  The expanded features are concatenated with the
encoder skip representation, linearly fused, and processed by two decoder
blocks with 12 heads.

Every fusion block contains two forms of interaction.  First, Earth-aware
window attention and an MLP mix spatial tokens independently inside each latent
slot.  Second, multi-head self-attention and an MLP mix the four latent slots at
each fixed geographic location.  If $\mathcal{B}_{s}$ and $\mathcal{B}_{z}$
denote these two operations, a fusion block is
\begin{equation}
\bm{Z}'=\mathcal{B}_{z}(\mathcal{B}_{s}(\bm{Z})).
\label{eq:app_fusion_block}
\end{equation}
This factorization separates multiscale spatial propagation from interaction
among latent semantics.

\subsubsection{Bounded Fusion-to-Field Feedback}

The fused latent memory is returned to the low-resolution field pathway by
cross-attention at each spatial location:
\begin{equation}
\begin{aligned}
\Delta\bm{X}^{l}=\operatorname{MHA}\big(&
Q=\operatorname{LN}(\bm{X}^{l}),
&K=\operatorname{LN}(\bm{Z}),
V=\operatorname{LN}(\bm{Z})\big).
\end{aligned}
\label{eq:app_feedback_delta}
\end{equation}
The amount written to each field and location is bounded:
\begin{equation}
\bm{\alpha}=0.25\,
\sigma\left(
f_{\alpha}([\bm{X}^{l},\Delta\bm{X}^{l},\bm{e}])
\right),
\label{eq:app_alpha}
\end{equation}
\begin{equation}
\overline{\bm{X}}^{l}=\bm{X}^{l}
+\bm{\alpha}\odot\Delta\bm{X}^{l}.
\label{eq:app_field_feedback}
\end{equation}
Thus $\alpha_{b,m,y,x}\in(0,0.25)$.  The field-dependent gate allows shared
multivariate information to be recovered selectively while preventing the
fusion pathway from overwhelming field-specific representations.

\subsection{Relation-Aware Forecast Decoder}

\subsubsection{Query Construction and Relation Prior}

For each registered output field $i$, the decoder constructs metadata using the
same variable, pressure, and surface parameterization as the encoder:
\begin{equation}
\bm{q}_i=\bm{E}_{\mathrm{var}}(v_i)
+\bm{E}_{\mathrm{pres}}(\widetilde p_i,s_i)
+\bm{E}_{\mathrm{surf}}(s_i).
\label{eq:app_query_metadata}
\end{equation}
A continuous embedding of the 6-hour transition interval is added at both
decoder resolutions.  The query is also conditioned on the local mean of the
available field memory:
\begin{equation}
\begin{aligned}
\overline{\bm{x}}_{b,y,x}
&=\frac{\sum_{m=1}^{M}P_{t,b,m}\bm{X}_{b,m,y,x}}
{\max(\sum_{m=1}^{M}P_{t,b,m},1)},\\
\bm{h}^{0}_{b,i,y,x}
&=\bm{q}_i+\overline{\bm{x}}_{b,y,x}.
\end{aligned}
\label{eq:app_contextual_query}
\end{equation}

The decoder uses the same relation registry as the dynamic graph, but translates
it into a fixed soft bias between query $i$ and input field $m$.  Let $r(i,m)$
denote their registered relation:
\begin{equation}
\eta_{i,m}=
\begin{cases}
2.0, & r(i,m)=\mathrm{self},\\
1.5-|\log p_i-\log p_m|, & r(i,m)=\mathrm{vertical},\\
0.75, & r(i,m)\in\mathcal{R}_{\mathrm{physical}}^{\ast},\\
-2.0, & r(i,m)=\mathrm{none},
\end{cases}
\label{eq:app_query_prior}
\end{equation}
where $\mathcal{R}_{\mathrm{physical}}^{\ast}$ contains all registered physical
relations other than self and vertical.
This is a preference rather than a hard mask; an unrelated but available field
can still be read if its learned content score is sufficiently informative.

\subsubsection{Low- and High-Resolution Memory Reading}

At the low-resolution scale, query $i$ attends to the updated field memory
$\overline{\bm{X}}^{l}$ at every location.  For attention head $a$,
\begin{equation}
s^{l}_{b,y,x,a,i,m}=
\frac{
\left\langle W_q^a\operatorname{LN}(\bm{h}^{0}_{b,i,y,x}),
W_k^a\overline{\bm{X}}^{l}_{b,m,y,x}\right\rangle
}{\sqrt{d_a}}
+\eta_{i,m}.
\label{eq:app_low_query_score}
\end{equation}
Unavailable fields are masked before softmax.  The resulting field-derived query
$\bm{Q}^{l}_{\mathrm{field}}$ subsequently reads the four fusion latents:
\begin{equation}
\bm{Q}^{l}_{\mathrm{fusion}}=\operatorname{MHA}
(\bm{Q}^{l}_{\mathrm{field}},\bm{Z},\bm{Z}).
\label{eq:app_query_reads_fusion}
\end{equation}
Their contributions are combined by a query- and location-dependent bounded
gate,
\begin{equation}
\bm{\beta}=0.5\,
\sigma\left(f_{\beta}
([\bm{Q}^{l}_{\mathrm{field}},
\bm{Q}^{l}_{\mathrm{fusion}},\bm{q}])\right),
\label{eq:app_beta}
\end{equation}
\begin{equation}
\bm{Q}^{l}=(1-\bm{\beta})\odot\bm{Q}^{l}_{\mathrm{field}}
+\bm{\beta}\odot\bm{Q}^{l}_{\mathrm{fusion}},
\label{eq:app_query_mix}
\end{equation}
where $\beta_{b,i,y,x}\in(0,0.5)$.

The low-resolution queries are projected and bilinearly expanded to
$31\times60$.  In parallel, an independent high-resolution query-to-field
attention reads $\bm{X}^{h}$ using the same metadata, local context, source
mask, and relation bias.  The high-resolution field output and expanded
low-resolution query are concatenated and linearly fused.  Two Earth-aware
query spatial blocks, one non-shifted and one shifted, then model spatial
dependencies within every output field before patch recovery.

\subsubsection{Patch Recovery and Stable Residual Update}

Field patch recovery linearly projects each query token to the 16 scalar values
of its $4\times4$ output patch.  Let $\bm{U}_{\theta,v}$ be the raw output for
variable $v$.  It is first softly bounded and then transformed by a
variable-specific affine map:
\begin{equation}
s_v=1+0.25\tanh(a_v),
\qquad
b_v=0.25\tanh(c_v),
\label{eq:app_recovery_affine}
\end{equation}
\begin{equation}
\begin{aligned}
\bm{R}_{\theta,v}
&=s_v\operatorname{SC}_{8}(\bm{U}_{\theta,v})+b_v,
\operatorname{SC}_{L}(x)&=L\tanh(x/L).
\end{aligned}
\label{eq:app_recovered_residual}
\end{equation}
Pressure levels of the same variable share $(s_v,b_v)$.  For the default
69-field query set, the final forecast is the normalized-space residual update
\begin{equation}
\widehat{\bm{X}}_{t+6\mathrm{h}}=
\operatorname{SC}_{12}
\left(\bm{X}_t+0.25\bm{R}_{\theta}\right).
\label{eq:app_residual_forecast}
\end{equation}
The internal $\operatorname{SC}_{8}$ limits the decoded update, while the final
$\operatorname{SC}_{12}$ limits the recursively propagated state.  These
operations are numerical stabilization mechanisms and do not impose mass,
energy, or momentum conservation.

\subsection{utoregressive Training and Inference}

\subsubsection{Masked Latitude-Weighted Huber Objective}

The reported model is optimized with a latitude-weighted Huber forecast loss in
per-field normalized space.  Let $\mu_{b,m,h,w}^{(s)}$ indicate that the target
location is valid and participates in supervision.  In particular, pressure
levels located below the local surface pressure can be excluded.  With
\begin{equation}
\omega_h=\frac{\max(\cos\phi_h,0)}
{H^{-1}\sum_{h'=1}^{H}\max(\cos\phi_{h'},0)},
\end{equation}
the loss at rollout step $s$ is
\begin{equation}
\mathcal{L}^{(s)}=
\frac{
\sum_{b,m,h,w}\mu_{b,m,h,w}^{(s)}\omega_h
\mathcal{H}_{\kappa}(e_{b,m,h,w}^{(s)})
}{
\max\left(
\sum_{b,m,h,w}\mu_{b,m,h,w}^{(s)}\omega_h,1
\right)
},
\label{eq:app_step_loss}
\end{equation}
where $e^{(s)}=\widehat{\bm{X}}^{(s)}-\bm{X}^{(s)}$ and
\begin{equation}
\mathcal{H}_{\kappa}(e)=
\begin{cases}
\frac{1}{2}e^2, & |e|\leq\kappa,\\
\kappa(|e|-\frac{1}{2}\kappa), & |e|>\kappa,
\end{cases}
\qquad \kappa=2.
\label{eq:app_huber}
\end{equation}
The normalization of $\omega_h$ gives the latitude weights unit mean.  No
auxiliary reconstruction loss or equation-based physical regularizer is used in
the reported objective.  For an $S$-step rollout, the total objective is
\begin{equation}
\mathcal{L}_{\mathrm{train}}=
\frac{1}{S}\sum_{s=1}^{S}\mathcal{L}^{(s)}.
\label{eq:app_rollout_loss}
\end{equation}

\subsubsection{Progressive Rollout Curriculum}

Training proceeds from a one-step transition to longer self-generated input
sequences.  Stage 1 supervises one 6-hour step.  Stage 2 initializes model
weights from the best Stage-1 checkpoint and supervises two autoregressive
steps.  Stage 3 initializes from the best Stage-2 checkpoint and supervises four
steps.  The optimizer and learning-rate schedule are restarted at each stage.

Let $\bm{X}^{(0)}=\bm{X}_t$.  At rollout step $s$,
\begin{equation}
\widehat{\bm{X}}^{(s)}=
\mathcal{F}_{\theta}
\left(
\bm{X}^{(s-1)},\bm{S},\bm{P}^{(s-1)},
\bm{e}^{\mathrm{cal}}_{t+s\delta}
\right),
\qquad \delta=6\,\mathrm{h}.
\label{eq:app_rollout_transition}
\end{equation}
Every application is conditioned on the same base transition interval, while
the target-valid calendar condition advances with $s$.  Thus a 24-hour forecast
is produced by four applications of the 6-hour operator, not by a direct
24-hour transition.

Only the first step inherits the availability mask of the original analysis.
Since each model output contains all 69 fields,
\begin{equation}
\bm{P}^{(s)}=\bm{1},\qquad s\geq1.
\label{eq:app_rollout_mask}
\end{equation}
Teacher forcing is disabled in Stage 1.  In Stages 2 and 3, the next input is
selected independently for each sample with probability $p_{\mathrm{TF}}=0.25$:
\begin{equation}
\begin{aligned}
\bm{X}^{(s)}
&=\xi_s\bm{Y}^{(s)}
+(1-\xi_s)\operatorname{stopgrad}
(\widehat{\bm{X}}^{(s)}),
\xi_s&\sim\operatorname{Bernoulli}(p_{\mathrm{TF}}),
\end{aligned}
\label{eq:app_teacher_forcing}
\end{equation}
where $\bm{Y}^{(s)}$ is the ground-truth state.  The stop-gradient operation
implements detached rollout: later-step losses update the shared transition
operator at that step but do not backpropagate through earlier predicted input
states.  This reduces memory usage and stabilizes the transition from one-step
training to self-generated multi-step inputs.

At inference time, teacher forcing is removed and every predicted state becomes
the input of the next transition.  The field-availability mask is set to one
after the first step, and the calendar features are updated at each valid time.
This inference procedure is identical for all reported lead times from three to
fourteen days.

\section{Data and Preprocessing Details}
We used the WeatherBench2 ERA5 dataset in the 69-variable configuration. The dynamic fields are defined on a global $240 \times 121$ longitude--latitude grid and sampled at 6-hour intervals. The atmospheric component consists of five variables: geopotential height $z$, temperature $t$, specific humidity $q$, zonal wind $u$, and meridional wind $v$. Each of these variables is available at 13 pressure levels (50, 100, 150, 200, 250, 300, 400, 500, 600, 700, 850, 925, and 1000 hPa), yielding 65 upper-air fields in total. In addition, we include four surface variables: 2 m temperature (\texttt{t2m}), 10 m zonal wind (\texttt{u10}), 10 m meridional wind (\texttt{v10}), and mean sea-level pressure (\texttt{msl}), giving 69 dynamic fields overall.

We followed the data split used in our implementation: 1979--2017 for training, 2018 for validation, and 2020 for testing. Unless otherwise stated, all dynamic variables were standardized using statistics computed on the training split only. Static boundary information, including the land--sea mask and orography, was provided as additional conditioning input. Forecast lead times were expressed as multiples of 6 h, so the reported 24 h, 168 h, and 336 h results correspond to 4, 28, and 56 autoregressive steps, respectively.

\section{Experimental Details}

\subsection{Evaluation Metrics}

We adopt two standard metrics widely used in weather forecasting evaluation: the latitude-weighted root mean square error (RMSE) and the anomaly correlation coefficient (ACC). Let $\widehat{X}_{b,m,h,w}^{(s)}$ and $X_{b,m,h,w}^{(s)}$ denote the forecast and ground-truth values for sample $b$, field $m$, latitude index $h\in\{1,\ldots,H\}$, and longitude index $w\in\{1,\ldots,W\}$ at rollout step $s$, where $H=121$ and $W=240$ are the numbers of latitude and longitude grid cells, respectively. The latitude weight at index $h$ with corresponding latitude $\phi_h$ is
\begin{equation}
\omega_h = \frac{\cos(\phi_h)}{H^{-1}\sum_{h'=1}^{H}\cos(\phi_{h'})},
\label{eq:lat_weight_app}
\end{equation}
so that $\frac{1}{H}\sum_{h=1}^{H}\omega_h = 1$ and hence $\sum_{h=1}^{H}\omega_h = H$.

\paragraph{Root Mean Square Error.}
Let $e_{b,m,h,w}^{(s)} = \widehat{X}_{b,m,h,w}^{(s)}-X_{b,m,h,w}^{(s)}$ denote the per-sample forecast error of field $m$ at step $s$. The latitude-weighted RMSE is
\begin{equation}
\operatorname{RMSE}_{m}^{(s)} = \sqrt{\frac{\sum_{b,h,w}\omega_h\,(e_{b,m,h,w}^{(s)})^{2}}{\sum_{b,h,w}\omega_h}}.
\label{eq:rmse_app}
\end{equation}
Normalized RMSE is obtained by dividing the latitude-weighted RMSE of each field by the corresponding training-set standard deviation, and the reported average is computed uniformly across all 69 fields.

\paragraph{Anomaly Correlation Coefficient.}
ACC measures the anomaly correlation between forecast and observation. Let $\overline{X}_{m,h,w}$ be the climatological mean of field $m$ at location $(h,w)$, and define the forecast and ground-truth anomalies as
\begin{equation}
\begin{aligned}
\tilde{X}_{b,m,h,w}^{(s)} = \widehat{X}_{b,m,h,w}^{(s)} - \overline{X}_{m,h,w},
\quad
\tilde{Y}_{b,m,h,w}^{(s)} = X_{b,m,h,w}^{(s)} - \overline{X}_{m,h,w}.
\label{eq:anomaly_def_app}
\end{aligned}
\end{equation}
The latitude-weighted ACC is then
\begin{equation}
\operatorname{ACC}_{m}^{(s)} = \frac{S_{XY}}{\sqrt{S_{XX}\;S_{YY}}},
\label{eq:acc_app}
\end{equation}
where $S_{XY}=\sum_{b,h,w}\omega_h\,\tilde{X}_{b,m,h,w}^{(s)}\,\tilde{Y}_{b,m,h,w}^{(s)}$, \,
$S_{XX}=\sum_{b,h,w}\omega_h(\tilde{X}_{b,m,h,w}^{(s)})^{2}$, \,
$S_{YY}=\sum_{b,h,w}\omega_h(\tilde{Y}_{b,m,h,w}^{(s)})^{2}$.
An ACC of~1 indicates perfect agreement, while lower values reflect reduced anomaly correlation. Throughout this paper, lower RMSE and higher ACC indicate better forecast skill.

\subsection{Analysis of Learned Graph Structure and Fusion Gates}
To examine whether VeinCast learns physically meaningful cross-variable interactions, we analysed the internal dynamic variable graph and the two fusion gates after training. This analysis was performed on the trained checkpoint without changing model parameters. During autoregressive inference, we registered forward hooks to collect three quantities: the dynamic adjacency matrix, the fusion feedback gate $\alpha$, and the fusion query gate $\beta$.

The diagnostic was conducted on the 2020 test period using 160 samples selected to cover all four seasons. We evaluated three forecast lead times, 24, 168, and 336 hours, corresponding to short-, medium-, and long-range rollout states. For each selected sample and lead time, the learned adjacency matrix was averaged over spatial windows and then grouped by variable family. The 69 forecast fields were partitioned into geopotential height ($z$), temperature ($t$), specific humidity ($q$), zonal wind ($u$), meridional wind ($v$), and surface variables. Rows in the adjacency matrix denote target fields and columns denote source fields.

We further decomposed the adjacency mass according to the edge types defined by the model's physical relation registry. These include self connections, vertical connections across pressure levels, thermodynamic--moisture links between $t$ and $q$, moisture-transport links between $q$ and $u/v$, wind--pressure links between $z$ and $u/v$, surface--atmosphere links, and learned residual edges outside the predefined physical relation set. This decomposition tests whether the learned graph simply uses unconstrained residual links or instead allocates most of its probability mass to meteorologically interpretable relations.

The learned graph showed a structured and non-uniform interaction pattern. The strongest edge-type mass was assigned to vertical relations, with an average mass of approximately 0.46 across the analysed lead times. Learned residual edges also received a non-negligible mass of about 0.16, while self connections contributed about 0.11. The remaining physically defined cross-family relations each received smaller but consistent weights. This pattern indicates that the model relies primarily on vertical coupling across pressure levels, while retaining a residual channel to capture interactions not covered by the hand-specified physical relation types.

We also analysed the two fusion gates to understand how the latent fusion pathway contributes to different variable families. The feedback gate $\alpha$ controls the strength of the residual update from fusion latents back to field tokens,
\begin{equation}
    h_i' = h_i + \alpha_i \Delta_i ,
\end{equation}
where $h_i$ is the field token and $\Delta_i$ is the fusion-derived update. Larger $\alpha$ values were observed for wind and surface variables, with mean values around 0.23, whereas $z$ and $q$ showed lower values around 0.16. This suggests that the recurrent state update uses the fusion pathway more strongly for dynamically coupled wind and surface fields.

The query gate $\beta$ controls how much the decoder query reads from the fusion representation,
\begin{equation}
    q_i' = (1-\beta_i) q_i + \beta_i \tilde{q}_i ,
\end{equation}
where $q_i$ is the field-specific query and $\tilde{q}_i$ is the fusion-query representation. The query gate showed a clearer separation across variable families. The largest values appeared for wind variables, especially $v$ and $u$, with mean $\beta$ values of approximately 0.47 and 0.44, respectively. In contrast, geopotential height had a much smaller mean value of approximately 0.08. These results indicate that the decoder relies more heavily on the global fusion pathway for wind-field readout, while geopotential fields are decoded mainly through field-specific representations.

Overall, this analysis shows that VeinCast does not use its graph and fusion modules as uniform black-box components. The learned adjacency emphasizes vertical atmospheric coupling, and the fusion gates allocate latent information differently across variable families. We interpret these observations as evidence that the model learns a structured internal representation aligned with major meteorological dependencies. However, this analysis is diagnostic rather than causal: the learned weights reveal how information is allocated inside the trained model, but they do not by themselves prove that each individual edge corresponds to a physical causal mechanism.

\section{Results Details}
\suppressfloats[t]
\subsection{RMSE and ACC of VeinCast and Baseline Models}

Table~\ref{tab:appendix_all_vars_1} \&~\ref{tab:appendix_all_vars_2} reports the latitude-weighted RMSE and ACC of VeinCast and the five baseline models on all remaining 300, 500, and 850\,hPa upper-air variables together with MSL, complementing the selected variables presented in the main text, and the evaluation covers lead times from 3 to 14 days. 

\begin{table*}[t!]
\caption{RMSE and ACC of VeinCast and baseline models on additional variables (Part~1: MSL, 300 hPa, and 500 hPa upper-air variables).}
\label{tab:appendix_all_vars_1}
\centering
\footnotesize
\renewcommand{\arraystretch}{1.0}
\setlength{\tabcolsep}{1.0mm}
\resizebox{\textwidth}{!}{%
\begin{tabular}{cc|cc|cc|cc|cc|cc|cc}
\toprule
\hline
\multicolumn{2}{c|}{Methods} & \multicolumn{2}{c|}{FengWu} & \multicolumn{2}{c|}{GraphCast} & \multicolumn{2}{c|}{Pangu-Weather} & \multicolumn{2}{c|}{ARROW} & \multicolumn{2}{c|}{FuXi} & \multicolumn{2}{c|}{VeinCast} \\
\hline
Variable & Lead Time & RMSE$\downarrow$ & ACC$\uparrow$ & RMSE$\downarrow$ & ACC$\uparrow$ & RMSE$\downarrow$ & ACC$\uparrow$ & RMSE$\downarrow$ & ACC$\uparrow$ & RMSE$\downarrow$ & ACC$\uparrow$ & RMSE$\downarrow$ & ACC$\uparrow$ \\
\hline
\hline
\multirow{5}{*}{MSL} & 3-day & 388.69 & 0.943 & 451.49 & 0.924 & 383.82 & 0.944 & 330.69 & 0.970 & 616.73 & 0.860 & 287.95 & 0.970 \\
 & 5-day & 605.58 & 0.851 & 722.68 & 0.800 & 556.69 & 0.881 & 580.99 & 0.904 & 877.57 & 0.732 & 504.17 & 0.910 \\
 & 7-day & 767.88 & 0.752 & 956.42 & 0.656 & 711.33 & 0.804 & 824.75 & 0.806 & 1034.71 & 0.641 & 702.64 & 0.828 \\
 & 10-day & 914.29 & 0.642 & 1189.21 & 0.502 & 860.20 & 0.713 & 1101.14 & 0.668 & 1173.20 & 0.552 & 894.67 & 0.724 \\
 & 14-day & 999.16 & 0.573 & 1345.62 & 0.424 & 950.34 & 0.652 & 1399.37 & 0.531 & 1247.11 & 0.521 & 1018.60 & 0.652 \\
\hline
\multirow{5}{*}{Z300} & 3-day & 515.46 & 0.994 & 406.63 & 0.996 & 473.88 & 0.995 & 413.04 & 0.996 & 706.80 & 0.987 & 381.84 & 0.996 \\
 & 5-day & 885.67 & 0.982 & 832.26 & 0.982 & 813.30 & 0.985 & 745.67 & 0.986 & 1140.95 & 0.967 & 700.63 & 0.987 \\
 & 7-day & 1203.96 & 0.968 & 1235.01 & 0.961 & 1099.64 & 0.972 & 1117.16 & 0.967 & 1426.98 & 0.949 & 1009.65 & 0.973 \\
 & 10-day & 1524.87 & 0.951 & 1662.72 & 0.934 & 1382.95 & 0.955 & 1675.85 & 0.925 & 1678.41 & 0.931 & 1323.07 & 0.953 \\
 & 14-day & 1771.88 & 0.939 & 1944.76 & 0.914 & 1564.40 & 0.942 & 2394.88 & 0.847 & 1822.30 & 0.919 & 1527.72 & 0.937 \\
\hline
\multirow{5}{*}{T300} & 3-day & 1.58 & 0.986 & 1.39 & 0.990 & 1.95 & 0.984 & 1.65 & 0.986 & 2.08 & 0.977 & 1.42 & 0.989 \\
 & 5-day & 2.25 & 0.973 & 2.27 & 0.974 & 2.76 & 0.968 & 2.41 & 0.970 & 3.01 & 0.954 & 2.00 & 0.979 \\
 & 7-day & 2.84 & 0.957 & 3.02 & 0.955 & 3.39 & 0.952 & 3.17 & 0.948 & 3.68 & 0.935 & 2.53 & 0.966 \\
 & 10-day & 3.45 & 0.938 & 3.82 & 0.930 & 3.99 & 0.932 & 4.32 & 0.904 & 4.32 & 0.914 & 3.12 & 0.949 \\
 & 14-day & 3.91 & 0.922 & 4.29 & 0.910 & 4.38 & 0.918 & 6.57 & 0.793 & 4.96 & 0.889 & 3.55 & 0.934 \\
\hline
\multirow{5}{*}{Q300} & 3-day & 1.20e-04 & 0.823 & 1.24e-04 & 0.838 & 1.20e-04 & 0.812 & 1.12e-04 & 0.841 & 1.46e-04 & 0.783 & 1.05e-04 & 0.859 \\
 & 5-day & 1.43e-04 & 0.749 & 1.65e-04 & 0.721 & 1.42e-04 & 0.731 & 1.39e-04 & 0.752 & 1.97e-04 & 0.644 & 1.26e-04 & 0.799 \\
 & 7-day & 1.57e-04 & 0.691 & 1.97e-04 & 0.609 & 1.56e-04 & 0.672 & 1.62e-04 & 0.664 & 2.34e-04 & 0.528 & 1.42e-04 & 0.745 \\
 & 10-day & 1.70e-04 & 0.633 & 2.27e-04 & 0.490 & 1.68e-04 & 0.612 & 1.92e-04 & 0.538 & 2.64e-04 & 0.432 & 1.60e-04 & 0.676 \\
 & 14-day & 1.77e-04 & 0.594 & 2.42e-04 & 0.402 & 1.74e-04 & 0.579 & 2.38e-04 & 0.376 & 2.84e-04 & 0.385 & 1.74e-04 & 0.617 \\
\hline
\multirow{5}{*}{U300} & 3-day & 6.53 & 0.929 & 6.08 & 0.939 & 6.89 & 0.921 & 6.43 & 0.935 & 8.84 & 0.870 & 6.08 & 0.938 \\
 & 5-day & 9.91 & 0.835 & 10.36 & 0.829 & 10.11 & 0.828 & 10.14 & 0.840 & 12.75 & 0.731 & 9.22 & 0.859 \\
 & 7-day & 12.58 & 0.735 & 13.73 & 0.706 & 12.49 & 0.736 & 13.31 & 0.727 & 15.15 & 0.621 & 11.98 & 0.765 \\
 & 10-day & 15.00 & 0.625 & 16.76 & 0.574 & 14.76 & 0.632 & 16.67 & 0.581 & 16.99 & 0.524 & 14.61 & 0.655 \\
 & 14-day & 16.57 & 0.548 & 18.53 & 0.487 & 16.08 & 0.566 & 19.56 & 0.450 & 18.28 & 0.459 & 16.33 & 0.577 \\
\hline
\multirow{5}{*}{V300} & 3-day & 7.09 & 0.837 & 6.26 & 0.873 & 7.08 & 0.833 & 6.40 & 0.870 & 9.52 & 0.709 & 6.35 & 0.869 \\
 & 5-day & 10.81 & 0.621 & 10.88 & 0.629 & 10.57 & 0.618 & 10.26 & 0.670 & 13.62 & 0.404 & 9.80 & 0.688 \\
 & 7-day & 13.61 & 0.397 & 14.38 & 0.373 & 13.21 & 0.399 & 13.50 & 0.441 & 15.79 & 0.193 & 12.79 & 0.472 \\
 & 10-day & 15.80 & 0.183 & 17.31 & 0.123 & 15.51 & 0.180 & 16.60 & 0.186 & 17.12 & 0.055 & 15.36 & 0.236 \\
 & 14-day & 16.83 & 0.071 & 18.34 & 0.033 & 16.52 & 0.077 & 18.36 & 0.069 & 17.52 & 0.014 & 16.73 & 0.099 \\
\hline
\multirow{5}{*}{Z500} & 3-day & 392.75 & 0.991 & 325.04 & 0.993 & 335.61 & 0.993 & 316.28 & 0.995 & 550.95 & 0.981 & 272.22 & 0.995 \\
 & 5-day & 663.65 & 0.976 & 629.48 & 0.975 & 573.01 & 0.980 & 560.84 & 0.982 & 866.36 & 0.955 & 506.64 & 0.984 \\
 & 7-day & 890.66 & 0.958 & 916.32 & 0.948 & 773.84 & 0.963 & 828.02 & 0.960 & 1069.90 & 0.933 & 733.05 & 0.966 \\
 & 10-day & 1112.65 & 0.936 & 1221.47 & 0.914 & 973.21 & 0.942 & 1212.46 & 0.912 & 1241.90 & 0.910 & 961.67 & 0.940 \\
 & 14-day & 1268.89 & 0.921 & 1430.54 & 0.891 & 1101.45 & 0.926 & 1671.97 & 0.839 & 1313.94 & 0.900 & 1109.36 & 0.920 \\
\hline
\multirow{5}{*}{T500} & 3-day & 1.82 & 0.986 & 1.54 & 0.990 & 1.83 & 0.988 & 1.76 & 0.988 & 2.37 & 0.977 & 1.53 & 0.990 \\
 & 5-day & 2.76 & 0.968 & 2.70 & 0.971 & 2.80 & 0.973 & 2.69 & 0.971 & 3.50 & 0.951 & 2.36 & 0.977 \\
 & 7-day & 3.53 & 0.949 & 3.69 & 0.948 & 3.58 & 0.956 & 3.58 & 0.948 & 4.21 & 0.930 & 3.12 & 0.959 \\
 & 10-day & 4.27 & 0.928 & 4.63 & 0.920 & 4.33 & 0.937 & 4.69 & 0.907 & 4.84 & 0.910 & 3.84 & 0.938 \\
 & 14-day & 4.78 & 0.913 & 5.18 & 0.900 & 4.79 & 0.925 & 6.49 & 0.822 & 5.30 & 0.892 & 4.30 & 0.922 \\
\hline
\multirow{5}{*}{U500} & 3-day & 4.65 & 0.923 & 4.49 & 0.930 & 4.76 & 0.919 & 4.49 & 0.931 & 6.35 & 0.861 & 4.30 & 0.935 \\
 & 5-day & 6.87 & 0.831 & 7.30 & 0.819 & 6.90 & 0.827 & 6.93 & 0.839 & 9.00 & 0.726 & 6.34 & 0.859 \\
 & 7-day & 8.63 & 0.731 & 9.58 & 0.698 & 8.50 & 0.735 & 9.11 & 0.726 & 10.62 & 0.622 & 8.16 & 0.768 \\
 & 10-day & 10.25 & 0.618 & 11.67 & 0.573 & 10.00 & 0.633 & 11.44 & 0.582 & 11.78 & 0.534 & 9.92 & 0.660 \\
 & 14-day & 11.26 & 0.541 & 12.82 & 0.506 & 10.89 & 0.569 & 13.46 & 0.465 & 12.45 & 0.488 & 11.04 & 0.583 \\
\hline
\multirow{5}{*}{V500} & 3-day & 4.98 & 0.822 & 4.56 & 0.852 & 4.83 & 0.826 & 4.47 & 0.861 & 6.76 & 0.679 & 4.42 & 0.858 \\
 & 5-day & 7.38 & 0.605 & 7.48 & 0.609 & 7.10 & 0.609 & 6.99 & 0.663 & 9.40 & 0.381 & 6.61 & 0.680 \\
 & 7-day & 9.15 & 0.388 & 9.73 & 0.361 & 8.81 & 0.394 & 9.12 & 0.436 & 10.78 & 0.183 & 8.51 & 0.470 \\
 & 10-day & 10.55 & 0.181 & 11.60 & 0.122 & 10.25 & 0.185 & 11.18 & 0.191 & 11.63 & 0.054 & 10.17 & 0.239 \\
 & 14-day & 11.22 & 0.076 & 12.29 & 0.031 & 10.96 & 0.082 & 12.52 & 0.073 & 11.83 & 0.013 & 11.05 & 0.105 \\
\hline
\bottomrule
\end{tabular}
}
\end{table*}

\begin{table*}[t!]
\caption{RMSE and ACC of VeinCast and baseline models on additional variables (Part~2: 850 hPa upper-air variables).}
\label{tab:appendix_all_vars_2}
\centering
\footnotesize
\renewcommand{\arraystretch}{1.0}
\setlength{\tabcolsep}{1.0mm}
\resizebox{\textwidth}{!}{%
\begin{tabular}{cc|cc|cc|cc|cc|cc|cc}
\toprule
\hline
\multicolumn{2}{c|}{Methods} & \multicolumn{2}{c|}{FengWu} & \multicolumn{2}{c|}{GraphCast} & \multicolumn{2}{c|}{Pangu-Weather} & \multicolumn{2}{c|}{ARROW} & \multicolumn{2}{c|}{FuXi} & \multicolumn{2}{c|}{VeinCast} \\
\hline
Variable & Lead Time & RMSE$\downarrow$ & ACC$\uparrow$ & RMSE$\downarrow$ & ACC$\uparrow$ & RMSE$\downarrow$ & ACC$\uparrow$ & RMSE$\downarrow$ & ACC$\uparrow$ & RMSE$\downarrow$ & ACC$\uparrow$ & RMSE$\downarrow$ & ACC$\uparrow$ \\
\hline
\hline
\multirow{5}{*}{Z850} & 3-day & 271.29 & 0.975 & 281.23 & 0.974 & 252.22 & 0.979 & 223.05 & 0.986 & 421.01 & 0.940 & 198.25 & 0.987 \\
 & 5-day & 435.30 & 0.935 & 492.76 & 0.917 & 405.66 & 0.942 & 397.27 & 0.953 & 629.53 & 0.872 & 362.82 & 0.956 \\
 & 7-day & 565.57 & 0.889 & 684.96 & 0.843 & 532.67 & 0.897 & 569.60 & 0.901 & 764.16 & 0.816 & 516.09 & 0.911 \\
 & 10-day & 683.74 & 0.835 & 885.88 & 0.759 & 655.61 & 0.840 & 761.82 & 0.822 & 877.65 & 0.761 & 667.23 & 0.850 \\
 & 14-day & 753.93 & 0.799 & 1031.77 & 0.714 & 728.14 & 0.801 & 968.36 & 0.730 & 921.55 & 0.744 & 764.08 & 0.805 \\
\hline
\multirow{5}{*}{Q850} & 3-day & 1.60e-03 & 0.927 & 1.74e-03 & 0.922 & 1.56e-03 & 0.931 & 1.64e-03 & 0.923 & 1.81e-03 & 0.905 & 1.43e-03 & 0.940 \\
 & 5-day & 1.91e-03 & 0.896 & 2.30e-03 & 0.869 & 1.91e-03 & 0.895 & 2.14e-03 & 0.871 & 2.26e-03 & 0.854 & 1.69e-03 & 0.915 \\
 & 7-day & 2.12e-03 & 0.871 & 2.70e-03 & 0.829 & 2.13e-03 & 0.869 & 2.58e-03 & 0.816 & 2.52e-03 & 0.819 & 1.91e-03 & 0.891 \\
 & 10-day & 2.31e-03 & 0.847 & 3.08e-03 & 0.789 & 2.32e-03 & 0.845 & 3.17e-03 & 0.733 & 2.78e-03 & 0.784 & 2.14e-03 & 0.864 \\
 & 14-day & 2.44e-03 & 0.831 & 3.28e-03 & 0.762 & 2.44e-03 & 0.829 & 3.98e-03 & 0.612 & 3.02e-03 & 0.747 & 2.31e-03 & 0.841 \\
\hline
\multirow{5}{*}{U850} & 3-day & 3.59 & 0.904 & 3.92 & 0.893 & 3.57 & 0.905 & 3.45 & 0.917 & 5.02 & 0.821 & 3.35 & 0.918 \\
 & 5-day & 5.07 & 0.806 & 5.93 & 0.760 & 5.01 & 0.807 & 5.22 & 0.812 & 6.85 & 0.680 & 4.75 & 0.836 \\
 & 7-day & 6.16 & 0.711 & 7.46 & 0.632 & 6.04 & 0.715 & 6.74 & 0.689 & 7.90 & 0.586 & 5.93 & 0.746 \\
 & 10-day & 7.05 & 0.614 & 8.70 & 0.515 & 6.91 & 0.623 & 8.29 & 0.539 & 8.69 & 0.507 & 6.99 & 0.647 \\
 & 14-day & 7.54 & 0.555 & 9.39 & 0.455 & 7.39 & 0.569 & 9.80 & 0.406 & 9.13 & 0.471 & 7.64 & 0.578 \\
\hline
\multirow{5}{*}{V850} & 3-day & 3.69 & 0.792 & 3.79 & 0.786 & 3.53 & 0.802 & 3.32 & 0.840 & 4.98 & 0.626 & 3.34 & 0.828 \\
 & 5-day & 5.20 & 0.577 & 5.67 & 0.533 & 4.96 & 0.589 & 5.07 & 0.634 & 6.55 & 0.346 & 4.75 & 0.649 \\
 & 7-day & 6.27 & 0.377 & 7.04 & 0.303 & 5.96 & 0.396 & 6.47 & 0.415 & 7.32 & 0.176 & 5.91 & 0.455 \\
 & 10-day & 7.05 & 0.203 & 8.09 & 0.109 & 6.78 & 0.221 & 7.72 & 0.205 & 7.82 & 0.070 & 6.87 & 0.262 \\
 & 14-day & 7.42 & 0.123 & 8.48 & 0.036 & 7.17 & 0.137 & 8.67 & 0.096 & 7.93 & 0.036 & 7.38 & 0.151 \\
\hline
\bottomrule
\end{tabular}
}
\end{table*}


\subsection{Visualization of VeinCast and Baseline Models}

\begin{figure*}[t!]
\centering
\includegraphics[width=\textwidth]{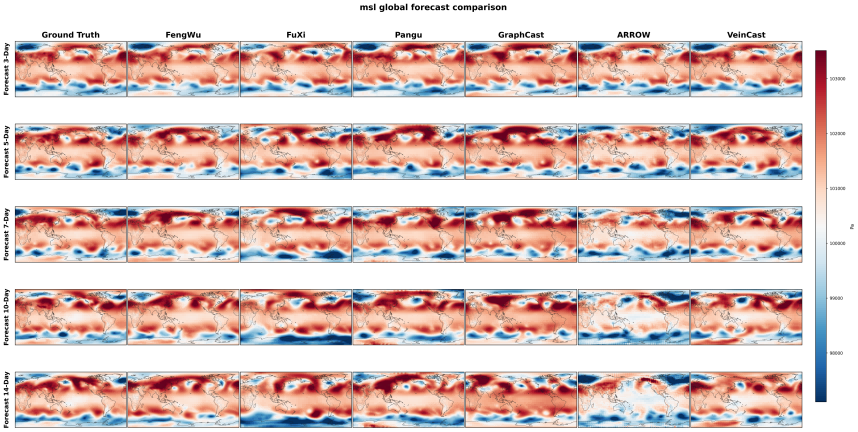}
\caption{Global forecast maps of VeinCast and baseline models for MSL (mean sea level pressure).}
\label{fig:app_msl_forecast}
\end{figure*}

\begin{figure*}[t!]
\centering
\includegraphics[width=\textwidth]{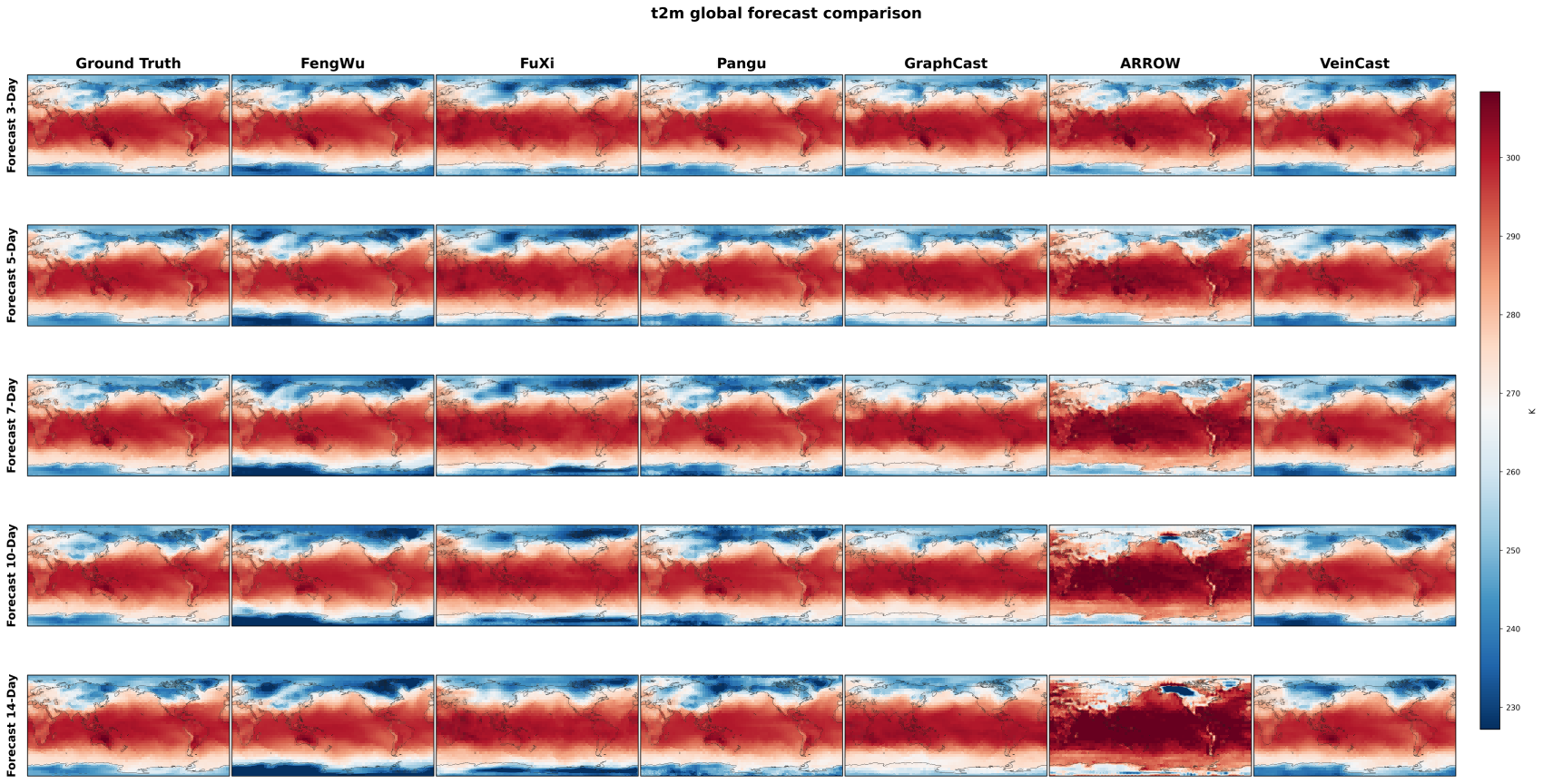}
\caption{Global forecast maps of VeinCast and baseline models for T2M (2\,m temperature).}
\label{fig:app_t2m_forecast}
\end{figure*}

\begin{figure*}[t!]
\centering
\includegraphics[width=\textwidth]{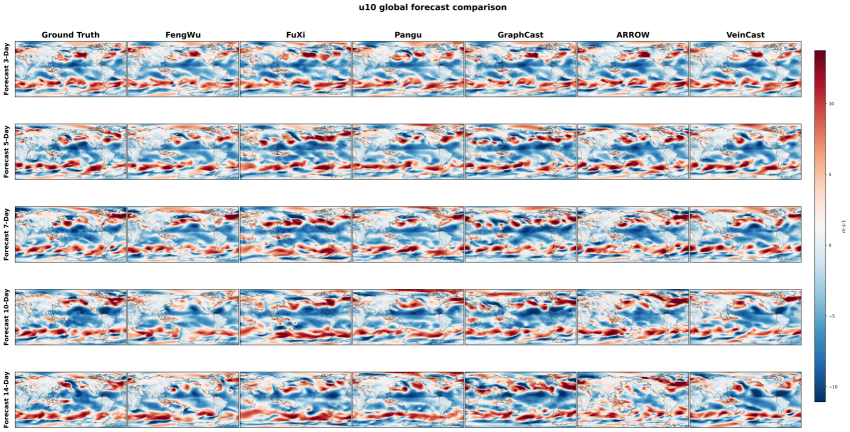}
\caption{Global forecast maps of VeinCast and baseline models for U10 (10\,m zonal wind).}
\label{fig:app_u10_forecast}
\end{figure*}

\begin{figure*}[t!]
\centering
\includegraphics[width=\textwidth]{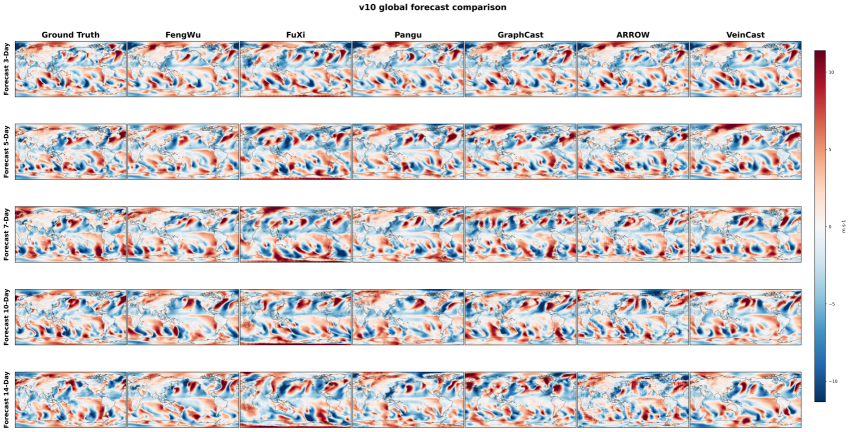}
\caption{Global forecast maps of VeinCast and baseline models for V10 (10\,m meridional wind).}
\label{fig:app_v10_forecast}
\end{figure*}

\begin{figure*}[t!]
\centering
\includegraphics[width=\textwidth]{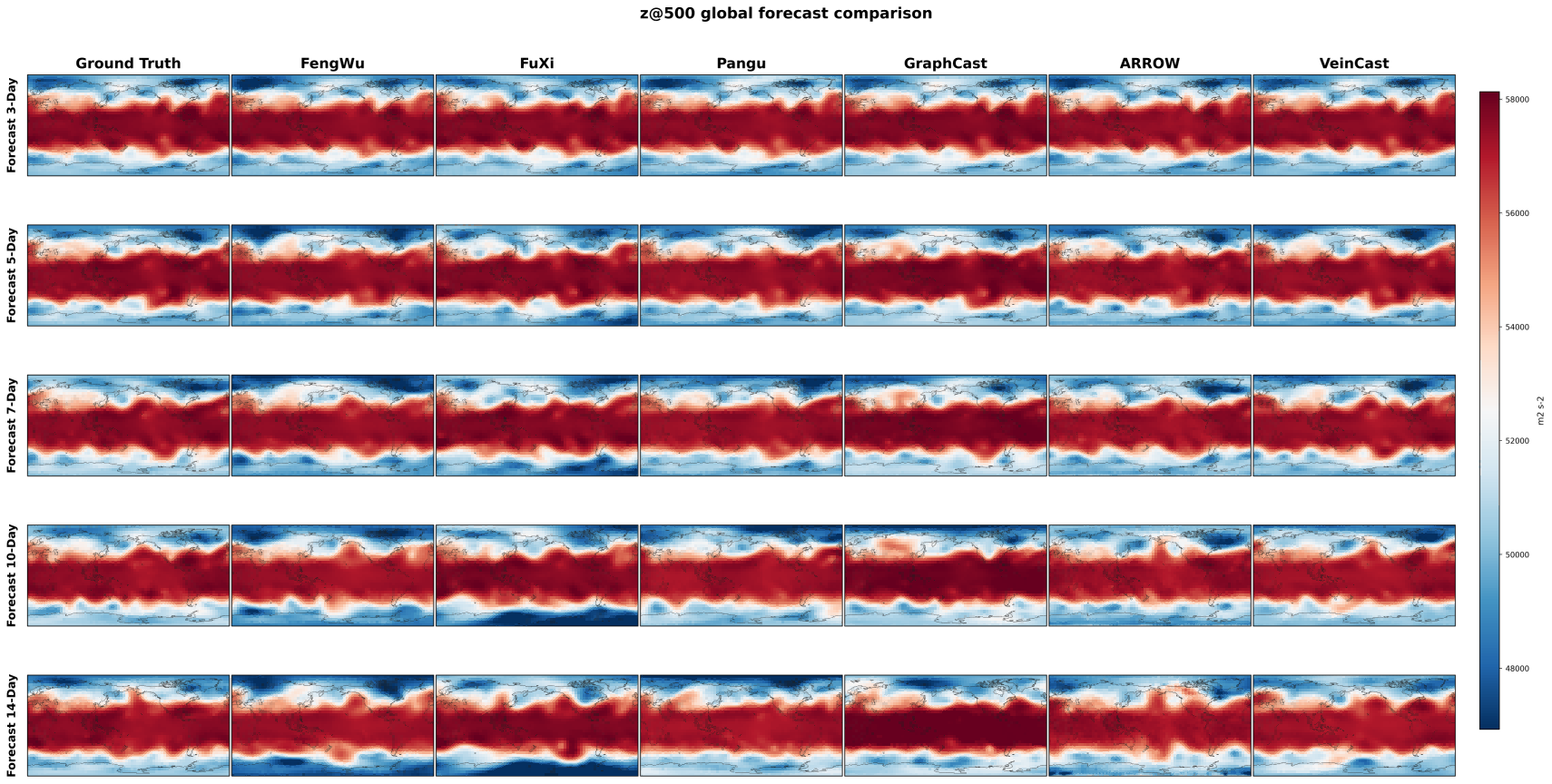}
\caption{Global forecast maps of VeinCast and baseline models for Z500 (geopotential at 500\,hPa).}
\label{fig:app_z500_forecast}
\end{figure*}

\begin{figure*}[t!]
\centering
\includegraphics[width=\textwidth]{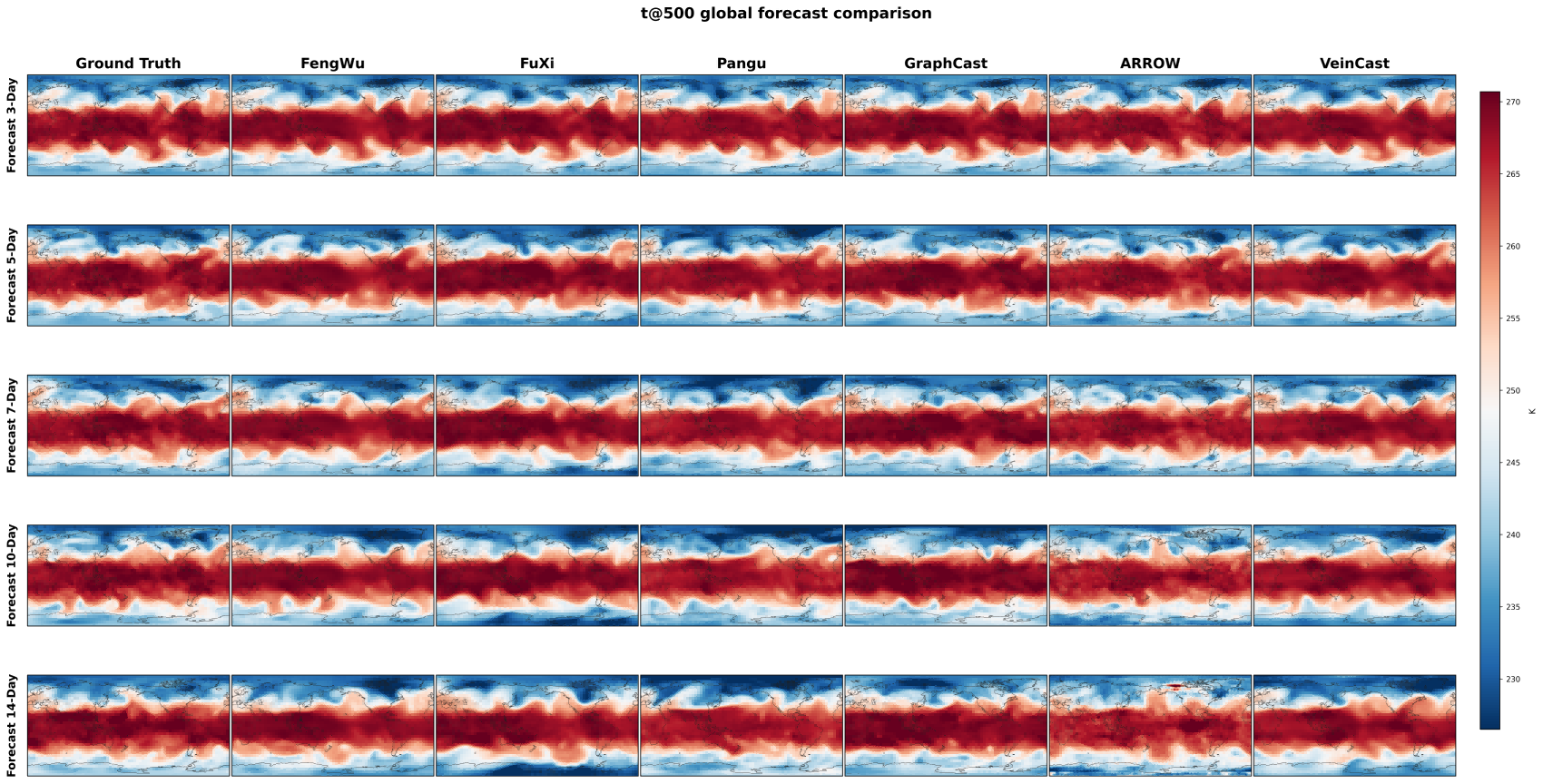}
\caption{Global forecast maps of VeinCast and baseline models for T500 (temperature at 500\,hPa).}
\label{fig:app_t500_forecast}
\end{figure*}

\begin{figure*}[t!]
\centering
\includegraphics[width=\textwidth]{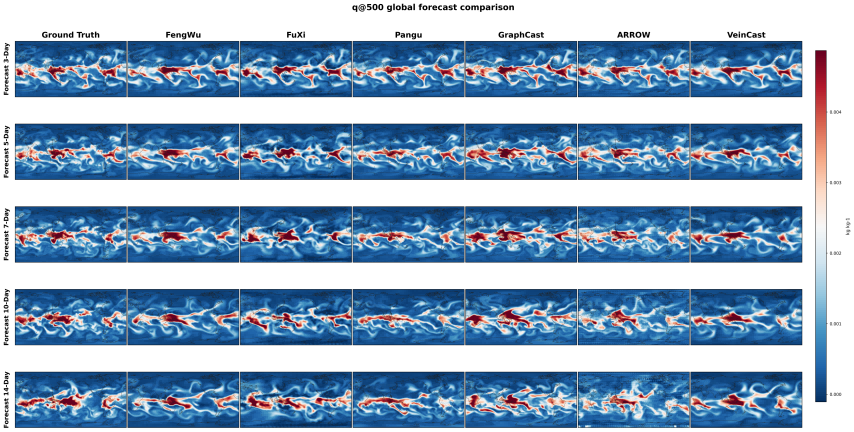}
\caption{Global forecast maps of VeinCast and baseline models for Q500 (specific humidity at 500\,hPa).}
\label{fig:app_q500_forecast}
\end{figure*}

\begin{figure*}[t!]
\centering
\includegraphics[width=\textwidth]{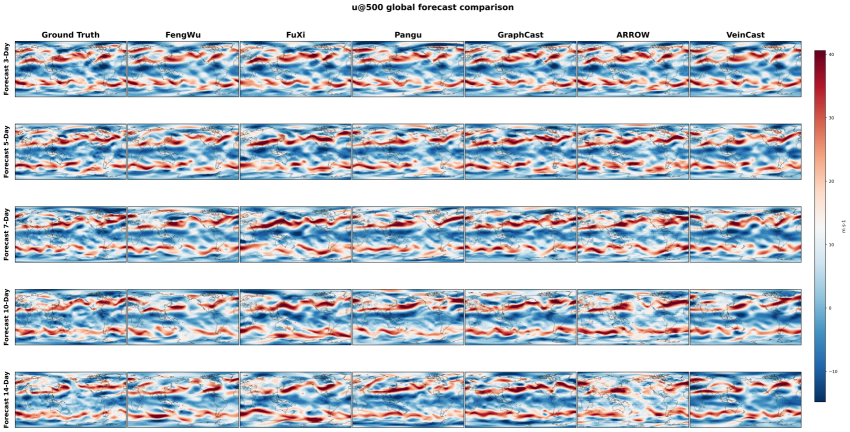}
\caption{Global forecast maps of VeinCast and baseline models for U500 (zonal wind at 500\,hPa).}
\label{fig:app_u500_forecast}
\end{figure*}

\begin{figure*}[t!]
\centering
\includegraphics[width=\textwidth]{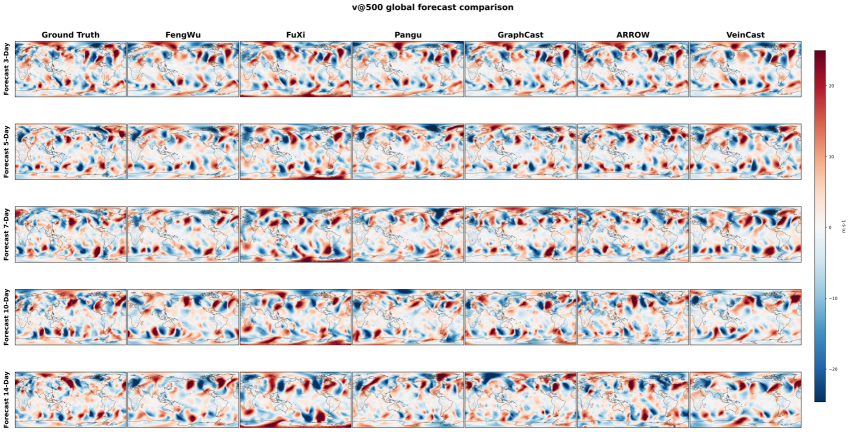}
\caption{Global forecast maps of VeinCast and baseline models for V500 (meridional wind at 500\,hPa).}
\label{fig:app_v500_forecast}
\end{figure*}

\begin{figure*}[t!]
\centering
\includegraphics[width=\textwidth]{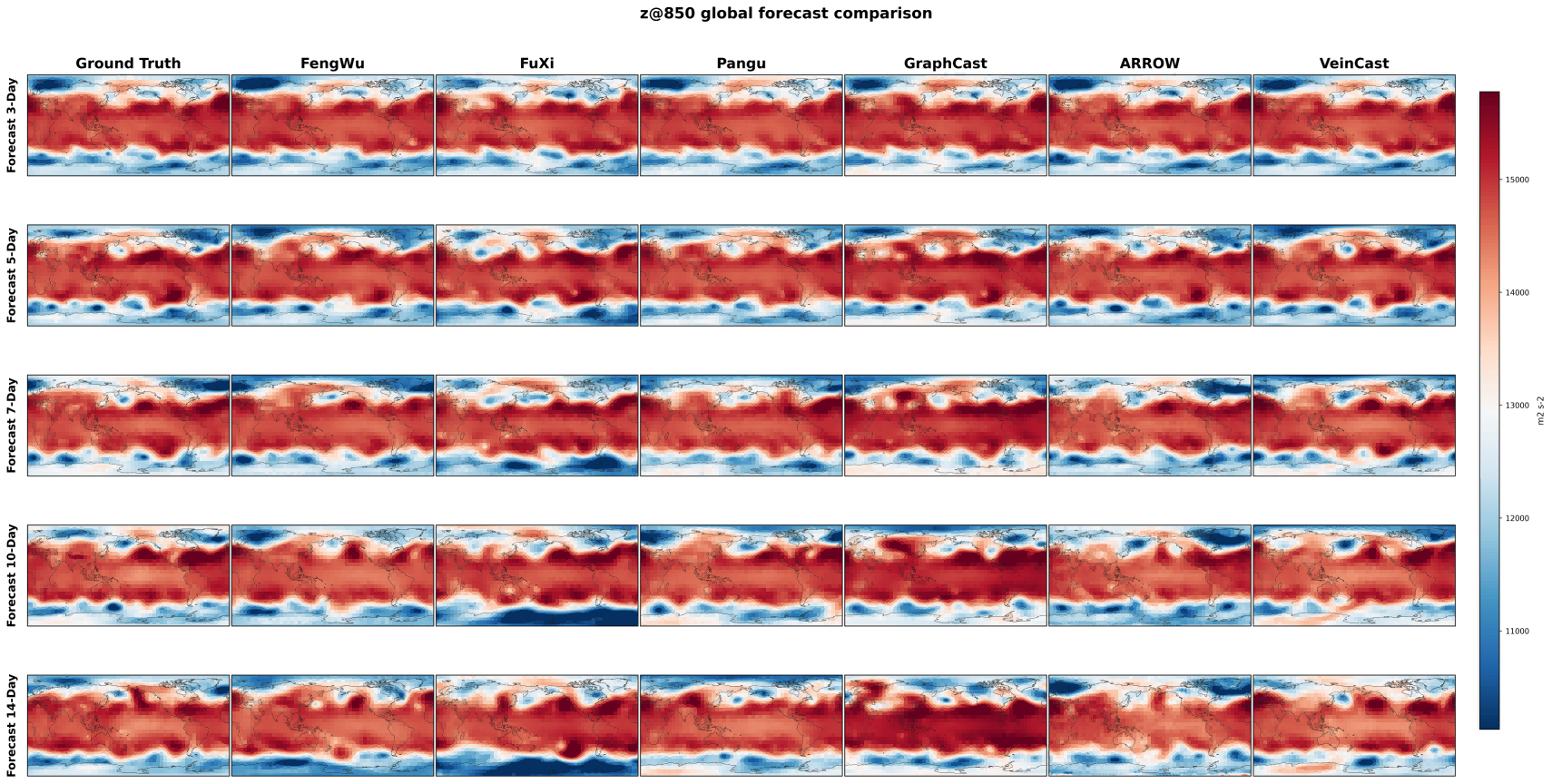}
\caption{Global forecast maps of VeinCast and baseline models for Z850 (geopotential at 850\,hPa).}
\label{fig:app_z850_forecast}
\end{figure*}

\begin{figure*}[t!]
\centering
\includegraphics[width=\textwidth]{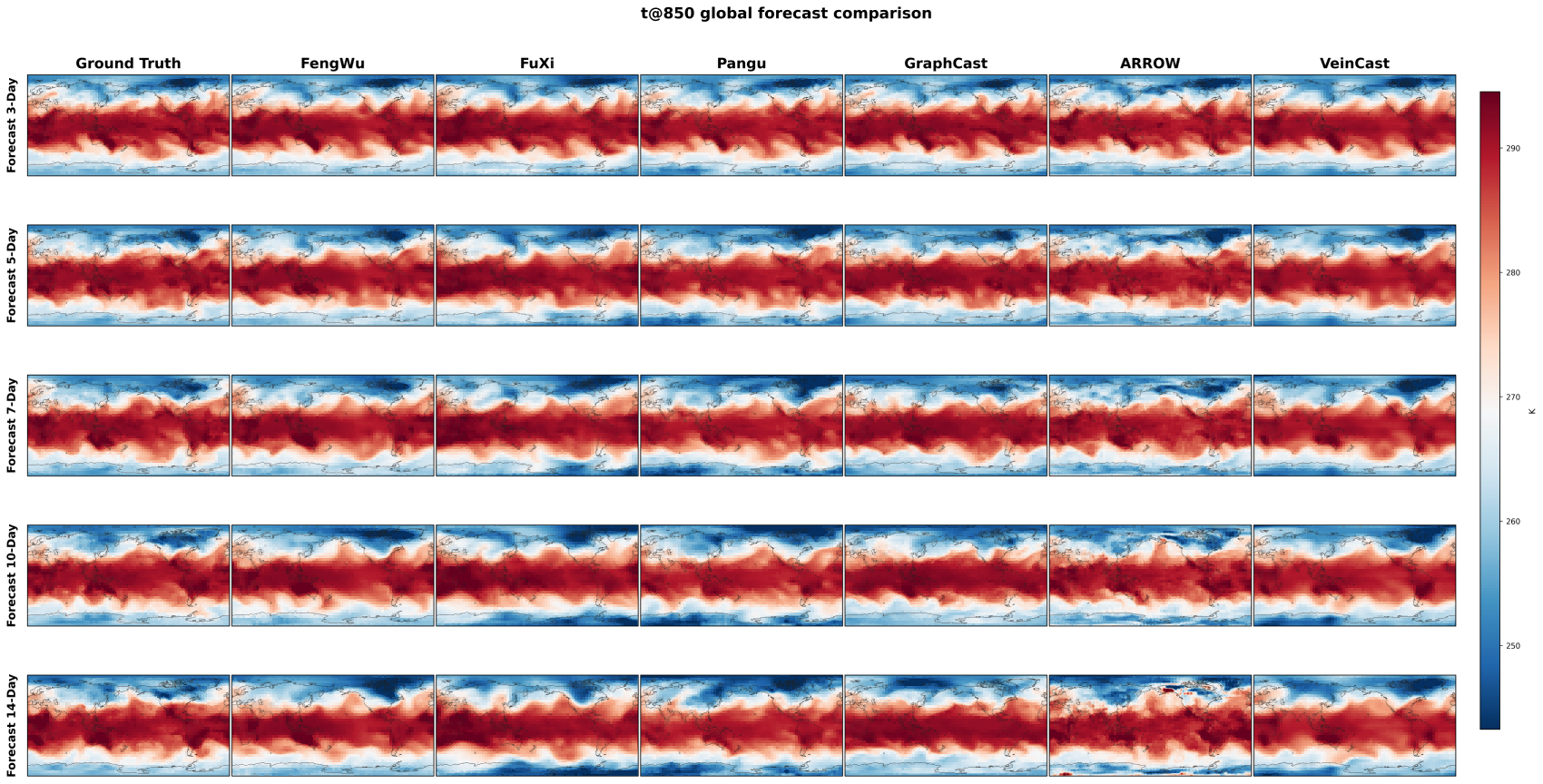}
\caption{Global forecast maps of VeinCast and baseline models for T850 (temperature at 850\,hPa).}
\label{fig:app_t850_forecast}
\end{figure*}

\begin{figure*}[t!]
\centering
\includegraphics[width=\textwidth]{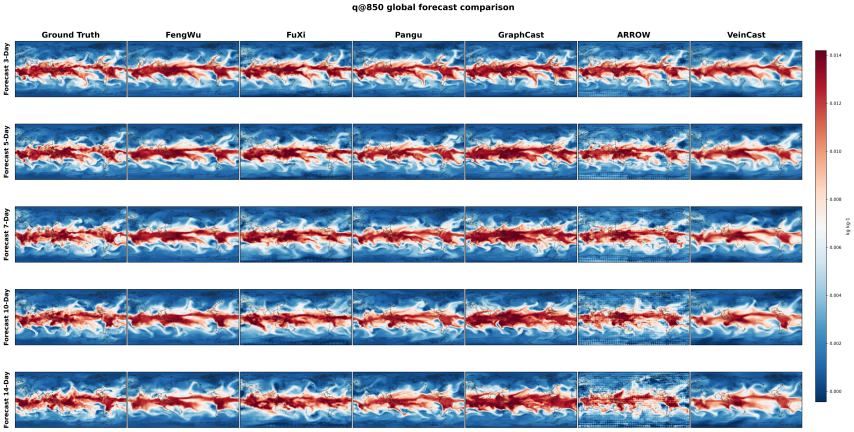}
\caption{Global forecast maps of VeinCast and baseline models for Q850 (specific humidity at 850\,hPa).}
\label{fig:app_q850_forecast}
\end{figure*}

\begin{figure*}[t!]
\centering
\includegraphics[width=\textwidth]{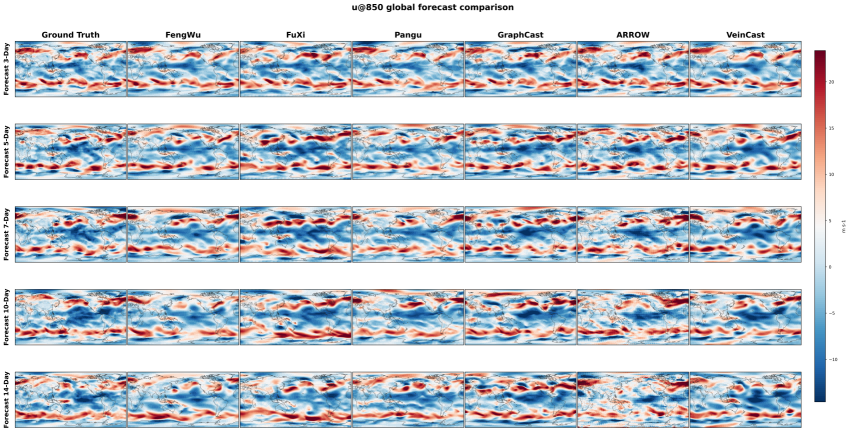}
\caption{Global forecast maps of VeinCast and baseline models for U850 (zonal wind at 850\,hPa).}
\label{fig:app_u850_forecast}
\end{figure*}

\begin{figure*}[t!]
\centering
\includegraphics[width=\textwidth]{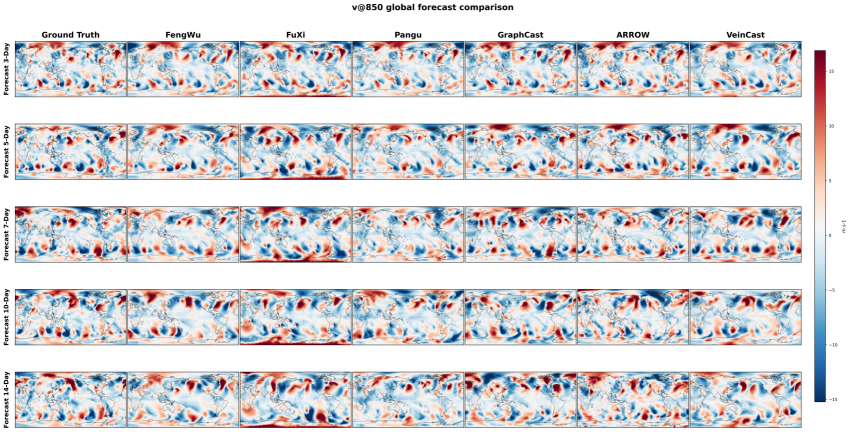}
\caption{Global forecast maps of VeinCast and baseline models for V850 (meridional wind at 850\,hPa).}
\label{fig:app_v850_forecast}
\end{figure*}

Figures~\ref{fig:app_msl_forecast}--\ref{fig:app_v850_forecast} provide global forecast maps for surface variables (MSL, T2M, U10, V10) and the five upper-air variables at 500\,hPa and 850\,hPa (Z, T, Q, U, V). For each variable, the ground-truth ERA5 field is shown together with the forecasts of VeinCast and the five baseline models at a representative lead time. These visualizations complement the quantitative results reported in the main text and Table~\ref{tab:appendix_all_vars_1} \&~\ref{tab:appendix_all_vars_2}.

\end{document}